\documentclass[11pt]{article}
\usepackage[a4paper, total={6in, 8in}]{geometry}
\usepackage{graphicx}
\usepackage{siunitx}
\usepackage{tikz}
\usepackage{float}
\usepackage{pifont}
\usepackage{wrapfig}  

\usepackage{amsmath,amsfonts,bm}

\def\eqref#1{equation~\ref{#1}}

\def\1{\bm{1}}

\def\rvmu{{\mathbf{\mu}}}
\def\rvtau{{\mathbf{\tau}}}

\def\vtheta{{\bm{\theta}}}

\def\vF{{\bm{F}}}

\def\vp{{\bm{p}}}
\def\vq{{\bm{q}}}

\def\vx{{\bm{x}}}
\def\vy{{\bm{y}}}

\makeatletter
\@ifundefined{mathsfit}{%
	\DeclareMathAlphabet{\mathsfit}{\encodingdefault}{\sfdefault}{m}{sl}%
	\SetMathAlphabet{\mathsfit}{bold}{\encodingdefault}{\sfdefault}{bx}{n}%
}{}
\makeatother

\usepackage{authblk} 
\usepackage{hyperref} 
\usepackage[finalnew]{trackchanges}
\usetikzlibrary{positioning,calc}

\usepackage[square,numbers]{natbib}
\begin{document}

\title{SSPINNpose: A Self-Supervised PINN for Inertial Pose and Dynamics Estimation}


\author[1]{Markus Gambietz}
\author[2]{Eva Dorschky}
\author[2]{Altan Akat}
\author[2]{Marcel Schöckel}
\author[3]{Jörg Miehling}
\author[1]{Anne D. Koelewijn}

\affil[1]{Chair of Autonomous Systems and Mechatronics, FAU Erlangen-Nürnberg, Erlangen, Germany}
\affil[2]{Machine Learning and Data Analytics Lab, FAU Erlangen-Nürnberg, Erlangen, Germany}
\affil[3]{Engineering Design, FAU Erlangen-Nürnberg, Erlangen, Germany}

\date{} 

\maketitle

\begin{abstract}
   Accurate real-time estimation of human movement dynamics, including internal joint moments and muscle forces, is essential for applications in clinical diagnostics and sports performance monitoring. Inertial measurement units (IMUs) provide a minimally intrusive solution for capturing motion data, particularly when used in sparse sensor configurations.
  However, current real-time methods rely on supervised learning, where a ground truth dataset needs to be measured with laboratory measurement systems, such as optical motion capture. 
  These systems are known to introduce measurement and processing errors and often fail to generalize to real-world or previously unseen movements, necessitating new data collection efforts that are time-consuming and impractical.
  To overcome these limitations, we propose SSPINNpose, a self-supervised, physics-informed neural network that estimates joint kinematics  and kinetics directly from IMU data, without requiring ground truth labels for training.
  We run the network output through a physics model of the human body to optimize physical plausibility and generate virtual measurement data. Using this virtual sensor data, the network is trained directly on the measured sensor data instead of a ground truth. When compared to optical motion capture, SSPINNpose is able to accurately estimate joint angles and joint moments at an RMSD of \SI{9.1}{\degree} and \SI{3.8}{{BWBH\%}}, respectively, for walking and running at  speeds up to \SI{4.9}{\meter\per\second} at a latency of \SI{3.5}{\milli\second}. Furthermore, the framework demonstrates robustness across sparse sensor configurations and can infer the anatomical locations of the sensors. These results underscore the potential of SSPINNpose as a scalable and adaptable solution for real-time biomechanical analysis in both laboratory and field environments.
\end{abstract}




\begin{figure*}[t]
  \includegraphics[width=\textwidth]{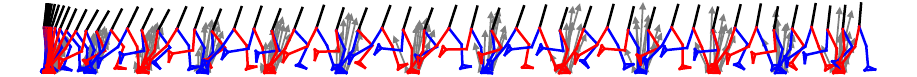}
  \caption{Example stickfigure of a running bout with a maximum speed of \SI{4.9}{\meter\per\second} reconstructed with SSPINNpose. We show the stick figure (black/red/blue) at intervals of \SI{100}{\milli\second} and the estimated GRFs (gray) every \SI{20}{\milli\second}.}
   \label{fig:intro}
\end{figure*}

\maketitle

\section{Introduction}
Understanding the injury mechanisms important for their prevention. However, injuries seldomly occur in controlled environments \cite{wallbankBicepsFemorisMuscle2024,heiderscheitIdentifyingTimeOccurrence2005}. Therefore, in-the-wild capturing of human movement dynamics, e.g. kinematics, joint torques, and ground reaction forces (GRFs), is required. 
Currently, the gold standard for capturing kinematics is optical motion capture (OMC), which is limited to a lab environment. In OMC, a person is fitted with reflective markers that are tracked by multiple cameras. 
Joint torques are estimated from the kinematics and force data, which are measured using force plates embedded into the floor, which further limits the environment. Applying the markers by hand is error-prone and the resulting kinematics can vary between different assessors and laboratories \cite{mcginleyReliabilityThreedimensionalKinematic2009a, fleischmannExploringDatasetBias2025}. Furthermore, different processing techniques can also lead to different results \cite{werlingRapidBilevelOptimization2022}. 

An alternative to OMC is the use of inertial measurement units (IMUs). These small, lightweight sensors can be worn during sports activities. Recent studies have explored methods that, based on inertial sensing, estimate poses \cite{yiTransPoseRealTime3D2021,vanwouweDiffusionPoserRealtimeHuman2024,vonmarcardSparseInertialPoser2017,huangDeepInertialPoser2018,jiangTransformerInertialPoser2022,roetenbergXsensMVNFull2013}, forces \cite{tanSelfSupervisedLearningImproves2024} or full dynamics \cite{karatsidisMusculoskeletalModelbasedInverse2019,dorschkyEstimationGaitKinematics2019,dorschkyCNNBasedEstimationSagittal2020,yiPhysicalInertialPoser2022,liReconstructingWalkingDynamics2021,winklerQuestSimHumanMotion2022}. The dynamics estimations are either based on deep learning \cite{yiTransPoseRealTime3D2021,winklerQuestSimHumanMotion2022,dorschkyCNNBasedEstimationSagittal2020}, trajectory optimization \cite{dorschkyEstimationGaitKinematics2019,liReconstructingWalkingDynamics2021} or static optimization~\cite{karatsidisMusculoskeletalModelbasedInverse2019}.
Current deep learning methods rely on supervised learning, which requires labeled data for training and, therefore, inherit the limitations and biases \cite{fleischmannExploringDatasetBias2025} of OMC. As a practical example, motions like high-speed running or sprinting require a large recording area, and are absent in widely used public IMU datasets like DIP-IMU and TotalCapture \cite{huangDeepInertialPoser2018,trumbleTotalCapture3D2017}. Additionally, these datasets do not include force data.
On the other hand, optimization-based methods need no labeled data but are computationally expensive, therefore not capable of real-time inference. This furthermore makes them infeasible for analyzing dynamics over a long time period, which, for example, could be a running session leading to an injury. 
Both deep learning and optimization-based methods can handle sparse IMU configurations \cite{winklerQuestSimHumanMotion2022,yiTransPoseRealTime3D2021,liReconstructingWalkingDynamics2021,dorschkyComparingSparseInertial2025}, where not every body part is equipped with an IMU. This can make a system more practical for the user, but also makes the reconstruction of human movement dynamics even more challenging \cite{vonmarcardSparseInertialPoser2017}. Similar to optical markers, the placement of IMUs can introduce errors in the kinematics estimation. Therefore, inferring the sensor placement from the data can be highly beneficial.

To address these limitations, we introduce SSPINNpose, which combines the real-time inference of learning-based methods with the ability of optimization-based approaches to reconstruct motions without relying on labeled data. The core principle behind SSPINNpose, a self-supervised physics-informed learning method, is that if an estimated motion is physically correct and corresponds to the measured IMU data, it is likely to be the correct motion. During training, the network is therefore guided to generate physically plausible motions that align with IMU data through virtual sensors. We further exploit auxiliary assumptions to accelerate training, mitigate local minima or enforce known properties of human movement.
Our main contribution is to transform the trajectory optimization problems from Li et al. \cite{liReconstructingWalkingDynamics2021} and Dorschky et al. \cite{dorschkyEstimationGaitKinematics2019} into a self-supervised learning problem, which allows for real-time inference. We further demonstrate that our method can be used with sparse IMU configurations and to estimate the IMU placement. To our knowledge, SSPINNpose is the first real-time method for estimating biomechanical variables from inertial sensor data without labeled training data. An example of our model's output is shown in Figure \ref{fig:intro}.


\section{Related Work}
Our work focuses on gait analysis, specifically the estimation of human movement dynamics, including both kinematics and the internal/external forces acting on the body. Since most dynamic motion during straight walking or running occurs in the lower limbs, particularly in the sagittal (forward-upward) plane, we review works that either examine full-body motion or focus on this plane.

\paragraph{Deep learning for movement dynamics:}
In order to estimate the 3D pose of a person in real-time from sparse IMU configurations, Huang et al. \cite{huangDeepInertialPoser2018} proposed a deep learning-based method using a recurrent neural network (RNN). Subsequent work enhanced motion accuracy and allowed for flexible sensor configurations \cite{yiTransPoseRealTime3D2021,vanwouweDiffusionPoserRealtimeHuman2024,jiangTransformerInertialPoser2022,zhangDynamicInertialPoser2024}, \add{with recent works accounting for body shapes}~\cite{yinShapeawareInertialPoser2025} \add{and measurement uncertainty}~\cite{Kim_2025_ICCV}. Since visually plausible motion was prioritized in these early methods, physical correctness, such as accurate force estimation, became a significant next step. 
Therefore, Winkler et al. \cite{winklerQuestSimHumanMotion2022} trained reinforcement learning agents to control torque-driven multibody dynamics models in a physical simulator. 
Another approach, Physical Inertial Poser (PIP), \cite{yiPhysicalInertialPoser2022} introduced a physics module to create physically plausible motions. The physics module contains a proportional-derivative (PD) controller and a motion optimizer, which also yields joint torques and GRFs, but only the kinematics have been validated so far. 
Similar to PIP, two-stage inference methods, where kinematics are first estimated with a learned prior and then dynamically updated with a physics model, are established in the domain of video-based pose estimation \cite{shimadaPhysCapPhysicallyPlausible2020,xiePhysicsbasedHumanMotion2022a,kocabasPACEHumanCamera2023,yiEgoLocateRealtimeMotion2023a,yuanSimPoESimulatedCharacter2021}.

In biomechanical applications, reference data was often recorded with OMC in combination with force plates, from which joint angles and torques can be estimated via inverse kinematics and inverse dynamics. Supervised deep learning models have demonstrated to accurately predict these outcome variables from IMU data in a single inference step. Examples span a range of applications, such as gait analysis \cite{limPredictionLowerLimb2019,hernandezLowerBodyKinematics2021,dorschkyCNNBasedEstimationSagittal2020}, slopes and stair climing \cite{chenDeterminingMotionsIMU2020}, activities of daily living \cite{wangEstimationLowerLimb2023} or pediatric care \cite{mohammadimoghadam3DGaitAnalysis2024}.

All deep learning-based inertial pose and dynamics estimation methods to date rely on labeled data for training. Therefore, these methods are unable to predict out-of-distribution movements. Furthermore, supervised methods inherit limitations from the reference system that was used for labelling, which is usually OMC, such as systemic and processing biases \cite{fleischmannExploringDatasetBias2025} or confinement to laboratory spaces.
Our method requires no labeled data for training as we use a fully self-supervised approach.

\paragraph{Optimization-based movement dynamics:}
To estimate movement dynamics without labeled data, one can use optimization-based methods. Based on kinematics estimated by the inertial motion capture system Xsens, Karatsidis et al- \cite{karatsidisMusculoskeletalModelbasedInverse2019} were first to propose the use of inverse methods to estimate GRFs and joint torques. From the estimated kinematics, they used static optimization to find  the GRF, and then used inverse dynamics to estimate the joint torques. They modeled the human body as a 3D musculoskeletal model with 39 degrees of freedom. However, their method has not been validated on running data and is not capable of handling sparse IMU setups or real-time inference. Furthermore, errors can accumulate during the multiple processing steps.

Movement dynamics can also be estimated in a single step with a trajectory optimization by finding control inputs, e.g. torques, for a simulation that best fits the IMU data. A solution to this can be found by solving an  optimal control problem. In optimal control, an objective function, in this case the difference between the actual and simulated IMU data, is minimized while satisfying dynamics constraints imposed by a multibody dynamics model. Dorscky et al. \cite{dorschkyEstimationGaitKinematics2019} solved the resulting optimization problem with a two-dimensional musculoskeletal model with 9 degrees of freedom and 7 IMUs using a direct collocation method. However, they assumed the gait to be symmetric and periodic. Furthermore, they only optimized on averaged gait cycles data from multiple trials, while inference took more than 30 minutes for a single gait cycle. They later followed up with a study on sparse IMU configurations under the same settings \cite{dorschkyComparingSparseInertial2025}. Optimal control problems with sparse IMU configurations under no symmetry assumptions have been solved by Li et al. \cite{liReconstructingWalkingDynamics2021}, but they relied on the detection of gait events instead. Detecting gait events from IMU data is an additional error source and unreliable for fast motions. 3D optimal control problems based on IMU data have \remove[2]{not been} \add[2]{recently been} solved \add[2]{for gait}~\cite{mcconnochieOptimalControlSimulations2025a}\change[2]{ yet, except}{, although complex maneuvers have only been simulated based on} \remove[2]{when }synthetic IMU data \remove[2]{was used} \cite{nitschke3DKinematicsKinetics2023}.

Our method is conceptually related to optimal control, as we aim to find a motion that minimizes the distance between actual and simulated IMU data and is physically plausible. Unlike optimal control, we create a surrogate model to stochastically map inputs to outputs instead of solving discrete optimization problems. A further difference is that optimal control problems use physical correctness as a constraint, while we use it as an optimization objective instead. This is similar to the solving strategy of constraint relaxation in optimization. As our method relies on stochastic optimization through a deep learning model, we use first-order solvers, such as Adam~\cite{kingmaAdamMethodStochastic2017}, instead of second-order solvers that are commonly used in optimal control problems, such as IPOPT \cite{wachterImplementationInteriorpointFilter2006}. Our method is advantageous in terms of pre-processing, as we do not need to detect gait events \cite{liReconstructingWalkingDynamics2021} or extract gait cycles under the assumption that these are periodic \cite{dorschkyEstimationGaitKinematics2019}.

\paragraph{From optimization problems to self-supervised learning:} 
Our work is based on the idea of transforming an optimization problem into a self-supervised learning problem. This paradigm leads to increased simulation speed, while not requiring labeled data and has been used in various fields. 
For example, for 3D human \cite{schmidtkeSelfsupervised3DHuman2023} and hand \cite{wanSelfSupervised3DHand2019} shape matching, the shape of a hand or human body was predicted from a single image with a neural network. The shape, was then (neurally) rendered and compared to the input image. As the rendering process is differentiable, they can backpropagate the error to the neural network. A similar approach was used for the design of RF pulses in MRI \cite{jangPhysicsguidedSelfsupervisedLearning2024}, where an optimal RF pulse prior was learned via MRI simulations. Self-supervised learning is also used in cloth simulation, where the neural network predicts the mechanics of clothing during movement, which is then evaluated based on physical plausibility \cite{bertichePBNSPhysicallyBased2021,santestebanSNUGSelfSupervisedNeural2022}. In SSPINNpose, we reconstruct our input signal, comparable to Wan et al. \cite{wanSelfSupervised3DHand2019}, and aim for physical plausibility as in Santesteban et al. \cite{santestebanSNUGSelfSupervisedNeural2022}.

\section{Method}
\subsection{Problem Formulation}
Our goal is to reconstruct lower body movement dynamics in the sagittal plane using IMUs. We aim to train a neural network for this task in a fully self-supervised manner, meaning that no labeled data for the outputs will be available during training.

\begin{figure}[h!]
  \begin{center}
  \includegraphics[width=\textwidth]{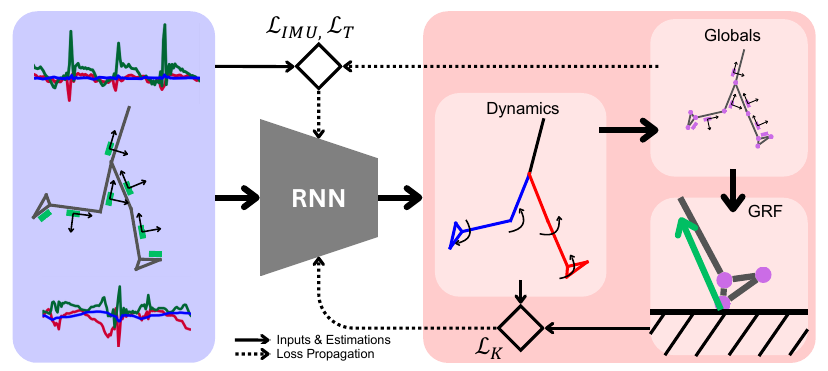}
  \end{center}
  \caption{Overview of the SSPINNpose's training scheme. The blue box shows inertial measure unit (IMU) signals from an unknown motion. For simplicity, we only show a single pose (gray). IMUs are annotated in light green. The RNN estimates the multibody dynamics in the first light red box. We then calculate the global kinematics for all joints, virtual IMUs, the heels and the toes (magenta). The ground reaction force (GRF, green) is then estimated based on the global ankle kinematics. Then we calculate the IMU loss ($\mathcal{L}_{IMU}$) and the temporal consistency loss ($\mathcal{L}_T$) based on the global positions and Kane's Loss ($\mathcal{L}_K$) based on the estimated joint angles, torques and GRFs.}
  \label{fig:model}
\end{figure}

The input consists of sequential two-dimensional accelerometer and gyroscope measurements from up to seven IMUs placed on the feet, shanks, thighs, and pelvis, alongside paramters that define the multibody dynamics model (Fig.~\ref{fig:model}). The outputs are the generalized coordinates, joint torques and GRFs. All outputs are directly estimated by the neural network, except for the GRFs, which are estimated with a ground contact model based on the kinematics output.

\subsection{SSPINNpose}\label{sec:sspinnpose}
We introduce SSPINNpose, a self-supervised physics-informed neural network designed to learn human movement dynamics from IMU data without labels. The term "physics-informed" refers to the integration of Kane's equations \cite{kaneDynamicsTheoryApplications1985} and a temporal consistency loss. Kane's equations ensure physical plausibility for all system states at each discritazation point, while the temporal consistency loss ensures that the estimated velocities and accelerations align with changes in position and velocity over time.
The self-supervised aspect relates to the reconstruction of the IMU data, allowing the model to learn from the inherent structure of the input signals. To ensure stable and fast training, we introduce further auxiliary losses that are based on either common assumptions in human movement or known properties of inertial sensors. In summary, SSPINNpose is trained with a weighted combination of the core (section~\ref{sec:core}) and auxiliary losses (section~\ref{sec:aux}, see Appendix~\ref{chap:implementationdetails} for more details \add{and Table}~\ref{tab:loss_functions} \add{for an overview)}:
\begin{equation}
   \mathcal{L} = \sum_{i \in \{IMU, T, K, GC\}} \lambda_{i} \mathcal{L}_{i} + \sum_{j \in \{B, \tau, slide, FS\}} \lambda_{j} \mathcal{L}_{j}.
\end{equation}

\begin{table*}[h!]
\centering
\caption{Overview of all loss functions used in SSPINNpose, divided into core loss functions (top four rows) and auxiliary loss functions (bottom four rows)}
\label{tab:loss_functions}
\begin{tabular}{lll}
\hline
\textbf{Loss Function} & \textbf{Name} & \textbf{Description} \\
\hline
$\mathcal{L}_K$ & {Kane's loss} & {Enforces physical plausibility per time step} \\
$\mathcal{L}_T$ & Temporal consistency & Enforces continuity of velocities and accelerations \\
$\mathcal{L}_{IMU}$ & IMU reconstruction & Tracks IMU data \\
$\mathcal{L}_{GC}$ & Ground contact & Realistic foot-ground interactions  \\
\hline
$\mathcal{L}_B$ & Bounds penalty & Joint limits and force bounds \\
$\mathcal{L}_{\tau}$ & Torque minimization & Effort minimization \\
$\mathcal{L}_{slide}$ & Sliding penalty & Prevents foot sliding \\
$\mathcal{L}_{FS}$ & Foot speed & Kalman filter-based constraint on foot speeds \\
\hline
\end{tabular}
\end{table*}

\subsubsection{RNN implementation}
To capture the temporal dependencies inherent in human movements and inertial sensor data, we employ a RNN. \change{We tested a LSTM [] for real-time inference and a bidirectional LSTM that has access to future information, each}{Concretely, we use a LSTM}~\cite{hochreiter1997long} followed by two dense layers to calculate the output. The network architecture is adapted from PIP~\cite{yiPhysicalInertialPoser2022}. At each time step $t$, the model receives the current IMU reading $\vx_t$, body constants $\vtheta_{b}$, IMU placement and rotations relative to their segment roots $\vtheta_{IMU}$, and ground contact model parameters $\vtheta_{gc}$. The input IMU data consists of 2D acceleration and 1D angular velocity data per sensor in the sagittal plane, and is augmented with Gaussian noise with a standard deviation of $\eta_{imu} \sigma(\vx_i)$ for each input channel $i$, where $\eta_{imu}$ is set to \SI{0.25}..

\begin{table}[hb!]
\centering
\caption{Notation used throughout the method description}
\label{tab:notation}
\begin{tabular}{ll}
\hline
\textbf{Symbol} & \textbf{Description} \\
\hline
\multicolumn{2}{l}{\textit{Network Inputs}} \\
$\vx_t$ & IMU reading at time step $t$ (accelerations and angular velocities) \\

$\vtheta_b$ & Body constants (mass, length, center of mass, moment of inertia) \\
$\vtheta_{IMU}$ & IMU placement and rotations relative to segment roots \\
$\vtheta_{gc}$ & Ground contact model parameters \\
\hline
\multicolumn{2}{l}{\textit{Network Outputs}} \\
$\hat{\vy}_t$ & Network output features at time step $t$ \\
$\vq, \dot{\vq}, \ddot{\vq}$ & Generalized coordinates, velocities, and accelerations \\
$\rvtau$ & Joint torques \\
$\tilde{\vq}_{ankle}, \tilde{\dot{\vq}}_{ankle}$ & Estimated global ankle kinematics \\
$\hat{\rvmu}$ & Learned current friction coefficient \\
\hline
\multicolumn{2}{l}{\textit{Calculated from Network Outputs}} \\
$\vp$ & Global kinematics: position $(x,y)$, angle $\alpha$, and derivatives \\
$\vF_{gc}$ & Ground reaction force \\
$\hat{\vx}_{imu}$ & Virtual IMU signals (simulated from estimated kinematics) \\
\hline
\end{tabular}
\end{table}

 The 46 output features $\hat{\vy}_t$ consist of each 9 estimated generalized coordinates $\vq$, velocities  $\dot{\vq}$, and accelerations $\ddot{\vq}$, 6 torques $\rvtau$ and 14 ground contact model states, which consists of the global kinematics of the ankle joint $\tilde{\vq}_{ankle}, \tilde{\dot{\vq}}_{ankle}$, and a current friction factor for each foot $\hat{\rvmu}$ \add{(see Table}~\ref{tab:notation} \add{for notation)}. The horizontal root position is ambiguous, therefore we replace it by the integrated horizontal root velocity.

For the loss calculations introduced in the following sections, we compute the global kinematics for the joints $\vp_j$, IMUs $\vp_{IMU}$ and ground contact points $\vp_{gc}$ based on the kinematics of the respective parent joint. The global kinematics of each point consist of its global position, $x,y$, and angle, $\alpha$, as well as their first and second derivatives $\vp = \{x,\dot{x},\ddot{x},y,\dot{y},\ddot{y},\alpha,\dot{\alpha},\ddot{\alpha}\}$ (see Appendix~\ref{chap:implementationdetails} for further details).

\subsubsection{Physics Information and Self-Supervision}\label{sec:core}
The main idea behind SSPINNpose is that a motion that is physically plausible and consistent with the IMU data is likely to be the correct motion. We enforce this by the following loss functions: Kane's loss ($\mathcal{L}_K$), temporal consistency loss ($\mathcal{L}_T$) and IMU reconstruction loss ($\mathcal{L}_{IMU}$). These core components of SSPINNpose are illustrated in Figure \ref{fig:model}.

\paragraph{Multibody Dynamics Model \& Kane's Equations:}
Our multibody dynamics model is a sagittal-plane lower limb model with 2 translational and 7 rotational degrees of freedom, which correspond to the generalized coordinates. The body consists of 7 segments: one trunk, and a thigh, shank, and foot for each leg. The body constants $\vtheta_{b}$ contain the mass, length, center of mass and moment of inertia for each segment. The body constants are linearly scaled based on the participant's height \cite{winterBiomechanicsMotorControl2009a}. We normalize all forces to bodyweight.

Using this dynamics model, we calculate the equations of motion based on Kane's method \cite{kaneDynamicsTheoryApplications1985}, implemented in SymPy \cite{meurerSymPySymbolicComputing2017}. Kane's formulation is advantageous for deep learning as it is the method that requires fewer equations to be solved to describe movement dynamics. Kane defined that the sum of internal ($ F_r^*$) and external ($F_r$) forces acting on a system is zero. Therefore, we can define a loss term that enforces the physical plausibility of each estimated state: 
\begin{equation}
   \mathcal{L}_K = \left(| \vF_r^* + \vF_r |\right)^2 = f\left(\hat{\vy}, \vtheta_{b}, \vF_{gc} \right)^2.
\end{equation}

To estimate the GRF $\vF_{gc}$, we model the foot-ground contact with a sliding contact point. The contact point's position between the heel and toe is determined based on the global ankle rotation. The vertical component of the GRF is modeled as a linear spring-damper system  as in \cite{vandenbogertImplicitMethodsEfficient2011a}, while the horizontal component is modeled as a friction cone with a learned current friction coefficient $\hat{\rvmu}$. To disentangle the GRF from the kinematics, we estimate the global ankle kinematics seperately, which is supervised by the difference with the estimated forward kinematics ($\mathcal{L}_{GC}$) of the ankle. For more details, see Appendix~\ref{chap:implementationdetails}. 

\paragraph{Temporal Consistency Loss:}
While Kane's method enforces physical plausibility at each time point, we also ensure that the derivatives of the estimated coordinates match the estimated velocities, and that the derivatives of the estimated velocities match the estimated accelerations. This loss is applied to the generalized coordinates $\vq$. We normalize the loss by the standard deviation of the estimated coordinates or velocities over the sequence to ensure scale-invariance:


\begin{equation}
   \mathcal{L}_T = \left(\frac{1}{2n_{\vq}} \sum_{i=1}^{n_\vq} \left(\left(\frac{d \vq_{i}} {d t} - \dot{\vq}_i\right) \sigma(\vq_i)^{-1} + \left(\frac{d \dot{\vq}_i} {d t} - \ddot{\vq}_i \right) \sigma(\dot{\vq}_i)^{-1} 
   \right)\right)^2.
\end{equation}

We chose this approximate integration method to decouple the learning of kinematics from movement dynamics, as numerical differentiation of the kinematics would cause exploding gradients in Kane's equations.

\paragraph{IMU Reconstruction Loss:}
We obtain virtual IMU signals $\hat{\vx}_{imu}$ \cite{dorschkyEstimationGaitKinematics2019} by rotating the kinematics of each IMU $\vp_{IMU}$ into its respective local coordinate system. These virtual IMU signals are then compared to the recorded IMU signals. We normalize the loss by the standard deviation over a sequence of the IMU signals per channel and the number of IMUs $n_{imu}$:
\begin{equation}
   \mathcal{L}_{IMU} = \left(\frac{1}{n_{imu}} \sum_{i=1}^{n_{imu}} \left( \vx_{imu} - \hat{\vx}_{imu} \right) \sigma(\vx_{imu})^{-1}\right)^2.
\end{equation}

\subsubsection{Auxiliary Losses}\label{sec:aux}
This section describes the auxiliary losses that we use to accelerate training, mitigate local minima or enforce known properties of human movement. For more details\change{ and an ablation a study}{, ablations and a sensitvity analysis} to justify these losses, refer to the Appendix sections \ref{chap:implementationdetails} and \ref{app:additionalresults}.
\paragraph{Joint Limit and Ground Contact Force Bounds $(\mathcal{L}_{B})$:} We penalize the model for exceeding joint limits and for violating bounds on maximum velocity and vertical position (see Appendix~\ref{chap:implementationdetails}). Additionally, we assume that for each sequence, each foot supports at least 20\% of the body weight. In practice, this avoids local minima where the model does not predict any ground contact or skips on one foot.
\paragraph{Torque Minimization ($\mathcal{L}_{\tau}$):} We apply a small weight on speed-weighted torque minimization, as minimizing effort is a common assumption in human movement and usually leads to more natural motions \cite{vandenbogertImplicitMethodsEfficient2011a}. Similar to \cite{dorschkyEstimationGaitKinematics2019}, we normalize the torques by the maximum speed of the root translation in the sagittal plane. As our training data might contain some non-movement phases, the speed normalization only applies to sequences with estimated moving speeds greater than \SI{1}{\meter\per\second}.
\paragraph{Sliding Penalty ($\mathcal{L}_{slide}$):}To prevent foot sliding when a ground reaction force (GRF) is present, we define sliding as the product of foot-ground speed and vertical GRF. This formulation ensures that at least one of these variables is constrained to be zero.
\paragraph{Foot Speed ($\mathcal{L}_{FS}$):} To speed up the training process and make our model less susceptible to local minima, we make use of known properties of foot-worn IMUs by reconstructing their global velocities ($\dot{\vp}_{K,x}$) using a Kalman filter with zero-velocity updates \cite{solaQuaternionKinematicsErrorstate2017,simoncolomarSmoothingZUPTaidedINSs2012}, as implemented in \cite{kuderleGaitmapOpenEcosystem2024}. In practice, this term contributes to avoiding local minima.
The IMU trajectory reconstruction algorithm is based on integration of the IMU signals which accumulates errors from drift and noise. Furthermore, zero-velocity updates are unreliable during running. As a consequence, we treat these reconstructed speeds as erroneous and only apply a penalty when the estimated foot-worn IMU speed from our kinematics differs by more than 30\% from its reconstructed maximum speed during the sequence.

\section{Experiments}\label{sec:experiments}
In this section, we first describe the dataset used for training and evaluation, followed by the evaluation metrics used to assess our model's performance. Next, we show and discuss model's capability to estimate human movement dynamics from IMU data in section \ref{baselines}. We then showcase that our model can be finetuned for more accurate force estimation and that, by optimizing our input variables, SSPINNpose is capable of estimating the IMU placement (section~\ref{physicsperso}). Finally, results and discussions on sparse IMU configurations (section \ref{sparsity}) are shown.

\paragraph{Dataset\add{s and Comparisons:}}
We use the "Lower-body Inertial Sensor and Optical Motion Capture Recordings of Walking and Running" dataset for training and evaluation \cite{dorschkyLowerbodyInertialSensor2024}. The dataset contains data of persons walking and running through an area equipped with OMC cameras and a single force plate, along with continuous IMU signals. For every trial, the OMC data contains roughly \SI{5}{\meter} of kinematics data and force plate data for a single step. We estimate the total running distance for the highest speed at ca. \SI{20}{\meter} based on SSPINNposes output. \remove{We downsampled the IMU signals to 100 Hz}. The dataset includes data from 10 participants, each performing 10 trials at 6 different speeds, ranging from \SI{0.9}{\meter\per\second} to \SI{4.9}{\meter\per\second}. For each condition, IMU data from the first 7 trials were designated for training, while the remaining 3 were used for evaluation. 

We selected the training data by applying a heuristic that identifies standing and turning phases based on the foot and pelvis IMU signals, respectively. This was done to include the run-up to the motion capture area and some steps after the motion capture area in our training set, while avoiding turning phases that we cannot reconstruct with a two-dimensional model. In total, our training data consists of 76 minutes of unlabeled IMU data. We processed the OMC and force plate data with \change{a}{A}ddBiomechanics \cite{werlingRapidBilevelOptimization2022} to compare the resulting joint angles and joint torques. The first participant was excluded from addBiomechanics because of erroneous force plate readings. 
During training, we randomly selected sequences of \SI{256} time steps from the training data, while full sequences were used during evaluation. Typical sequences from the datasets are visualized in Figures \ref{fig:intro} and Appendix Figure \ref{headliner}.
This dataset has been used by several other works focussing on sagittal-plane lower limb dynamics \cite{dorschkyEstimationGaitKinematics2019,dorschkyCNNBasedEstimationSagittal2020,dorschkyComparingSparseInertial2025}. \add[1]{Our preprocessing differs from prior work. As our model does not require labels or segmented gait cycles, we use continuous IMU sequences without extracting representative gait cycles} \cite{dorschkyCNNBasedEstimationSagittal2020} \add[1]{or averaging multiple gait cycles} \cite{dorschkyEstimationGaitKinematics2019,dorschkyComparingSparseInertial2025}. \add[1]{Thus, SSPINNpose is trained directly on raw, continuous motion data.}
\add{In additional experiments, we use the ensemble-averaged gait cycles from} \cite{dorschkyEstimationGaitKinematics2019} \add{to compare against their physics-based, label-free optimal control method. For this, we retrained our model randomly selected sequences of 256 time steps from their (circular-padded) ensemble-averaged gait cycles. 
Finally, we trained our model on the continuous IMU data from Molinaro et al.} \cite{molinaroDatasetTaskAgnosticExoskeleton2024}\add{, where we withheld three participants for validation. For this dataset, which contains a range of cyclic and non-cyclic motions, no reference speed data is available. All datasets were resampled to} \SI{100}{\hertz}.

\paragraph{Metrics:} We use the following metrics to evaluate our model: \textit{1.) Joint Angle Error (JAE):} The root mean square deviation (RMSD) between the estimated joint angles and those obtained from addBiomechanics, including the root orientation, in degrees.  \textit{2.) Joint Torque Error (JTE):} The RMSD between the estimated joint torques and those obtained from addBiomechanics, in bodyweight-bodyheight percent (\unit{{BWBH\%}}). \textit{3.) GRF Error (GRFE):} The root mean square error (RMSE) between the estimated GRFs and those obtained from the force plate, normalized by the bodyweight, in bodyweight percent (\unit{{BW\%}}). The GRF is the only outcome variable that can be directly measured, therefore, we consider it to be an error and not a deviation to a reference system. \textit{4.) Speed Error:} The RMSD between the estimated average speed and the sagittal-plane speed of the pelvis markers while the participant was crossing the OMC area, in \unit{\meter\per\second}. For all metrics, lower values are better. We show an evaluation on metrics that are commonly used in computer graphics in the Appendix \ref{app:3dpose}.

\subsection{Quantitative and Qualitative Evaluation} \label{baselines}

\begin{table}[hb!]
  \caption{Quantitative comparisons. We compare SSPINNpose when trained, and when finetuned (FT), on continuous data against a CNN-based regression method \cite{dorschkyCNNBasedEstimationSagittal2020} and against an physics-based optimal control method \cite{dorschkyEstimationGaitKinematics2019} on ensemble-averaged gait cycles. We also evaluate SSPINNpose on the Molinaro et al. dataset \cite{molinaroDatasetTaskAgnosticExoskeleton2024}, for which baseline comparisons are discussed in the text below due to different evaluation protocols per data modality. The metrics are defined in section \ref{sec:experiments}.} 
  \label{tab:baselineall}
  \begin{center}
  \begin{tabular}{l|lcccccc}
     \hline
     \textbf{Dataset} & \textbf{Model} & \textbf{JAE} & \textbf{JTE}& \textbf{GRFE}& \textbf{Speed} & \textbf{No labels} & \textbf{Latency} \\
     & & [deg] & [\unit{{BWBH\%}}]&  [\unit{{BW\%}}]& [\unit{\meter\per\second}] & \textbf{needed}\\
     \hline
     Dorschky \cite{dorschkyLowerbodyInertialSensor2024} & Ours & 9.1 & 3.8 & 16.6 & 0.15 & \ding{51} & \SI{3.5}{\milli\second} \\ 
     (continuous) & Ours (FT) & 9.4 & 3.3 & 13.3 & 0.22 & \ding{51} & \SI{3.5}{\milli\second} \\
     & CNN \cite{dorschkyCNNBasedEstimationSagittal2020} & 4.9 & 1.4 & 10.7 & - & \ding{55} & <\SI{1}{\milli\second}* \\
     \hline 
     Dorschky \cite{dorschkyLowerbodyInertialSensor2024} & Ours & 6.4 & 1.8 & 12.7 & 0.12 & \ding{51} & \SI{3.5}{\milli\second}\\
     (averaged) & OCP \cite{dorschkyEstimationGaitKinematics2019} & 6.3 & 2.6 & 17.9 & 0.25 & \ding{51} & \SI{50}{\minute}\\
     \hline
     Molinaro \cite{molinaroDatasetTaskAgnosticExoskeleton2024} & Ours & 11.0 & 1.8 & 16.2 & - & \ding{51} & \SI{3.5}{\milli\second} \\
     \hline
  \end{tabular}
  \begin{tabular}{r}
  \textbf{*}needs to wait for a full gait cycle to complete before inference. \\
   \end{tabular}
  \end{center}
\end{table}

\add{Training SSPINNpose on continuous IMU data from} \cite{dorschkyLowerbodyInertialSensor2024} \add{took approximately 6 hours on a NVIDIA RTX 4090 GPU. The resulting network was able to estimate joint angles, joint moments, ground reaction forces, and speed in real-time, with a latency of {3.5} {ms}. We then obtained a physics-finetuned version of the model (FT) by continouing the training with ten-fold increased weight on physics-based losses} ($\mathcal{L}_K$, $\mathcal{L}_T$ and $\mathcal{L}_{GC}$).
Compared to existing biomechanically validated methods, SSPINNpose is able to estimate the dynamics of human movement from IMU data in real-time without the need for labeled data. We achieve a speed error that is \SI{0.1}{\meter\per\second} smaller the current optimal control-based state of the art~\cite{dorschkyComparingSparseInertial2025}.

\begin{figure}[h!]
   \begin{center}
      \includegraphics{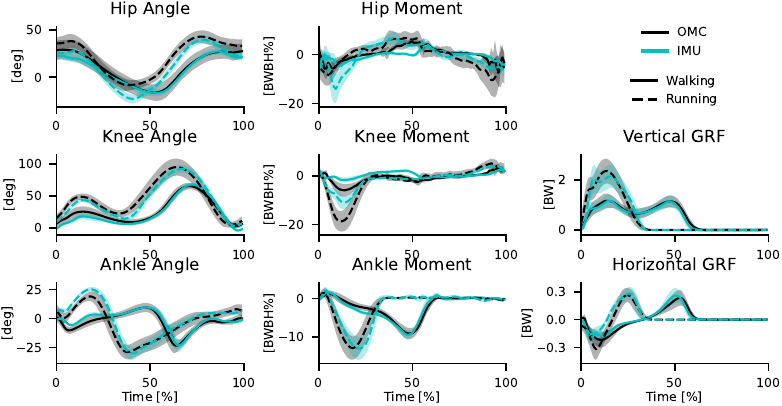}
   \end{center}
   \caption{Average joint angles, torques and ground reaction forces (GRFs) for the right leg over all test gait cycles\change{. Estimated with the Bi-LSTM.}{ estimated with the finetuned model.} We segmented the gait cycles during which the force plate was hit and normalized them to a duration of 100 samples. Walking and running data is shown in solid and dashed lines, respectively. Our estimates are shown in cyan, the reference data is shown in black. The shaded area represents the standard deviation.}
   \label{fig:JAJEGRF}
\end{figure}

\add{For the remaining metrics, we achieve a JAE of} \SI{9.1}{\degree}, a JTE of \SI{3.8}{\percent} BWBH, and a GRFE of \SI{16.6}{\percent} BW (Table \ref{tab:baselineall}). \add{The finetuned model trades of minor inaccuracies in speed and joint angle estimation for improved torque and GRF estimation, with a JTE of} \SI{3.3}{\percent} BWBH \add{and a GRFE of} \SI{13.7}{\percent} BW. \add{Here, SSPINNpose is outperformed by the supervised CNN-based regression method from} \cite{dorschkyCNNBasedEstimationSagittal2020}\add{, which however requires pre-segmented gait cycles as inputs and does not estimate gait speed. 
When trained on ensemble-averaged gait cycles, SSPINNpose achieves similar JAE to the optimal control-based state of the art}~\cite{dorschkyEstimationGaitKinematics2019}, \add{while having smaller JTE, GRFE and speed error}, see Table~\ref{tab:baselineall}.
Finally, we trained SSPINNpose on raw IMU data from Molinaro et al. \cite{molinaroDatasetTaskAgnosticExoskeleton2024} \add{and achieved similar performance} (Table~\ref{tab:baselineall}). \add{For this dataset, a supervised method achieved a 44\% lower JTE, but made use of additional pressure insole and encoder data} \cite{molinaroTaskagnosticExoskeletonControl2024a}. \add{As the dataset contains the largest amount of labeled data, we tested a supervised baseline that directly regresses the joint angles and ground reaction forces from the IMU data with the same LSTM architecture as SSPINNpose.} \add{This supervised baseline achieved a lower JAE than SSPINNpose} (\SI{7.7}{\degree}), \add{while the GRFE was slightly higher at} \SI{18.9}{{BW\%}}.

\remove{SSPINNPOSE was able to predict joint angles, joint moments, ground reaction forces, and speed with similar accuracy when using an LSTM and a Bi-LSTM (Table ). We evaluated  SSPINNpose's performance on continuous IMU data using a LSTM and a Bi-LSTM model, respectively. Between both, there are only minor differences in the outcome metrics. The LSTM model estimated dynamics and GRFs slightly more accurately, while the Bi-LSTM model estimated speed more accurately and produced smoother motions. The LSTM can estimate the joint angles and torques in real-time, with a latency of {3.5}{ms}. Training took approximately 16 hours on a NVIDIA RTX 3080 GPU. To compare against state of the art methods that report results on the same dataset, we show versions of our model with adapted training and evaluation schemes. To compare to [], which optimized on ensemble averaged gait cycles, we trained and evaluated SSPINNpose (OCP) on ensemble averaged gait cycles from all {60} trials. Here, the JTE, GRFE and speed error are slightly higher compared to training and evaluation on continuous data.}


\remove{SSPINNpose's kinematics estimations are on par with current real-time deep learning-based methods [] (see Appendix~ for more details).
The JAE, JTE, and GRFE, on the other hand, are generally larger than in regression-based methods such as ~[], when tested under the same conditions.}

In Figure \ref{fig:JAJEGRF}, we show the gait-cycle averages of the \add{estimated} joint angles, torques and GRFs \remove{estimated with the Bi-LSTM model }in comparison to the OMC reference. The kinematics were estimated accurately, with a small bias in the hip and knee angle. \change{Especially in running, the hip and knee moment were not accurately estimated during the stance phase, which is the first 40\% of the gait cycle for running and the first 60\% for walking. The ankle moment and vertical GRF shows slightly lower values than the reference data, while the horizontal GRF could not be estimated correctly}{Errors in joint moments and ground reaction force estimations are highest during the inital stance phase (the first 20\% of the gait cycles)}. SSPINNpose estimated the kinematics and speeds robustly, with median and 95th percentile errors of \SI{5.5}{\degree} and \SI{16.6}{\degree} for joint angles, and \SI{4.0}{\percent} and \SI{10.6}{\percent} for speed.


\subsection{Sparse IMU Configurations}\label{sparsity}
\begin{table}[t]
   \caption{Comparison of different sparse IMU configurations on the evaluation metrics, using the continuous IMU data from \cite{dorschkyLowerbodyInertialSensor2024}.}
   \label{tab:spawseIMU}
   \begin{center}
   \begin{tabular}{lccccc}
      \hline
      \textbf{IMU configuration} & \textbf{JAE} & \textbf{JTE} & \textbf{GRFE} & \textbf{Speed} \\
      & [deg] & [\unit{{BWBH\%}}]&  [\unit{{BW\%}}]& [\unit{\meter\per\second}] \\
      \hline
      All & 9.1 & 3.8 & 16.6 & 0.15 \\
      Feet + Thighs & 9.8 & 4.2 & 18.3 & 0.19 \\
      Feet + Shank & 10.5 & 4.0 & 17.1 & 0.27 \\
      Shank + Pelvis & 13.6 & 8.2 & 30.7 & 0.19 \\
      Shank + Thighs & 17.3 & 6.2 & 23.6 & 0.33 \\
      \hline
   \end{tabular}
   \end{center}
\end{table}

\begin{figure}[h!]
   \begin{center}
   \includegraphics[width=4in]{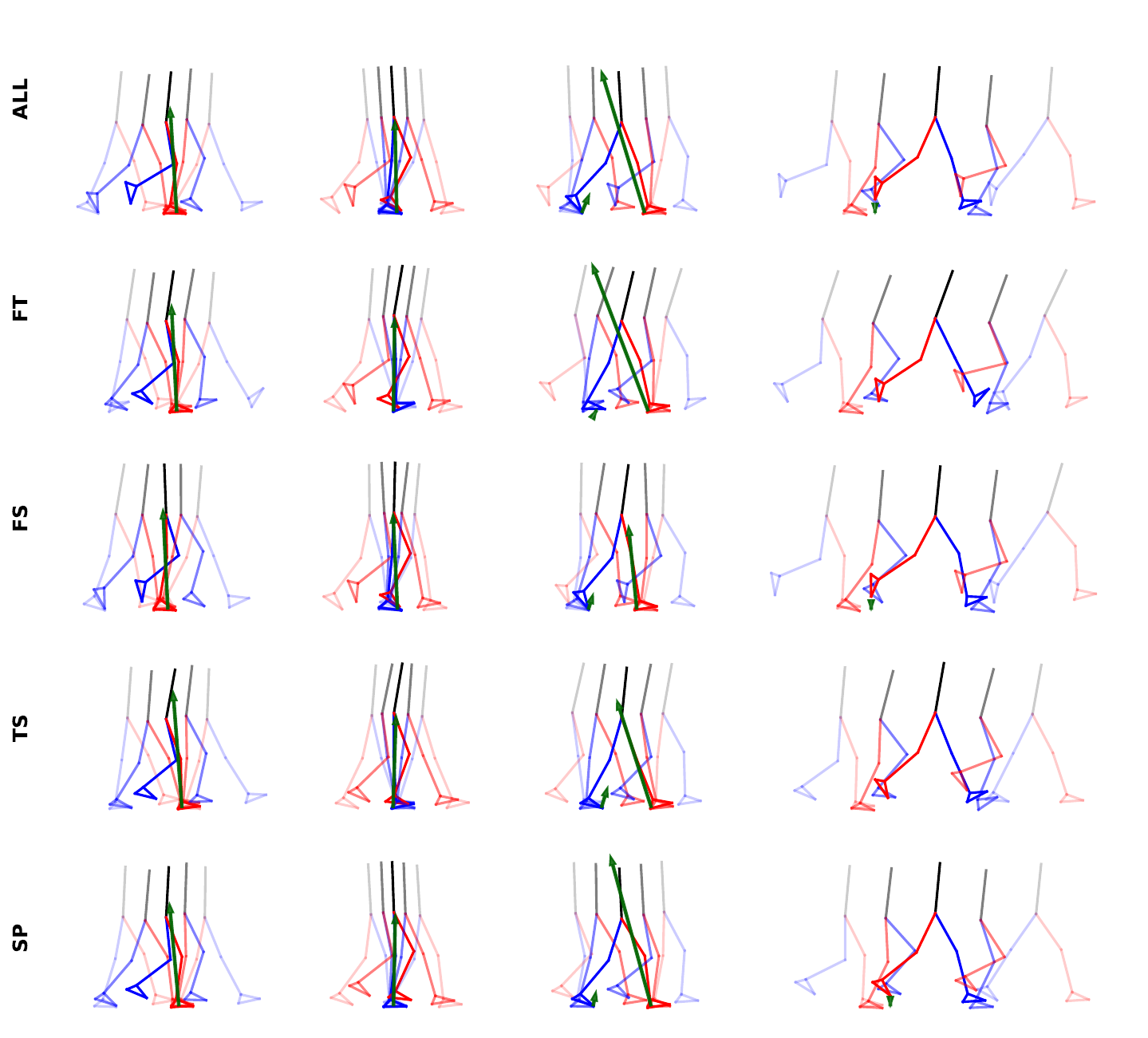}
   \end{center}
   \caption{Samples of stick figure sequences for sparse IMU configurations. Configurations shown are ALL (Baseline), FT (Foot + Thigh), FS (Foot + Shank), TS (Shank + Thigh), and SP (Shank + Pelvis). Each sequences shows 5 frames at different time points, where subsequent frames are shown at 150 milliseconds intervals. The middle frame has forces annotated in green.}
   \label{fig:spawsestick}
\end{figure}

\change{In practice, the fewer IMUs one has to wear, the better.}{Using fewer IMUs improves usability.} Therefore\add{, we} retrained \change{the Bi-LSTM}{SSPINNpose} from scratch on configurations 
with only the \remove{foot-worn IMUs (F),} foot and thigh IMUs (FT), foot and \change{pelvis}{shank} IMUs (FS), \add{shank and thigh IMUs (SP), and shank and pelvis IMUs (SP)}. Errors generally increased (Table \ref{tab:spawseIMU}), but the output motion is still physically and visually plausible (see Fig.~\ref{fig:spawsestick}). \change{For the running motions in F and FT configurations, the ankle angle and therefore the origin of the GRF is visibly shifted.} Between the configurations with and without a pelvis IMU, the trunk orientation is different for all motions. \add{As there are fewer IMUs, and therefore less motion information available, we increased the weight of the IMU reconstruction loss $\mathcal{L}_{IMU}$ so that the relative weight between the loss terms remained similar to the full IMU configuration. We furthermore tested configurations with only pelvis, thigh, shank, or foot IMUs, respectively, but these configurations did not lead to meaningful results.}

\subsection{IMU Positioning Optimization} \label{physicsperso}
\remove{Depending on the outcome variable of interest, we can finetune SSPINNpose by prioritizing different loss terms. In an ideal simulation, the estimated dynamics should perfectly match the actual motion. However, achieving a perfect simulation requires physical exactness, meaning that both Kane's loss and the temporal consistency loss must be zero. Therefore, we finetuned the LSTM towards physics by increasing the weight of the Kane's loss and the temporal consistency loss by a factor of 10. This reduced the JTE by 10\% and the GRFE by 20\%. However, as the IMU signals were not followed as strictly, the JAE increased by 5\% and the speed error increased by 33\%. After this finetuning, the biases in knee moment and vertical GRF were substantially reduced and only the bias in the hip torque during the stance phase in running remained. For use cases where the torques are of most interest, this trade-off should be acceptable.}

\begin{figure}[h!]
   \begin{center}
   \includegraphics{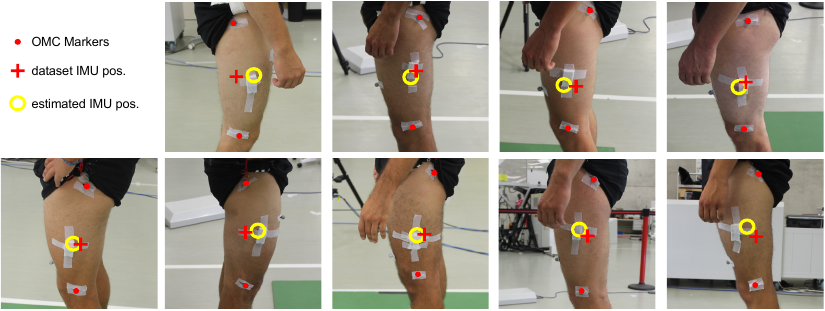}
   \end{center}
   \caption{Comparison of IMU positionings from the dataset and our estimations. We use OMC markers as a reference frame. For all participants, we show either the right or left leg. We always chose the side where the IMU and OMC markers were clearly visible. If they were visible from both sides, we chose the picture that was taken more perpendicular to the sagittal plane.}
   \label{fig:imuposthigh}
\end{figure}

A perfect simulation would require a correct multibody dynamics model with correct IMU positions. Our model and loss function \add{together }can act \remove{together }as a differential physical simulator. Therefore, we can optimize input parameters, including IMU orientations and positions. The IMU orientations and positions are prone to errors as they are placed and measured manually. Therefore, we finetuned the network and the IMU positions and orientations jointly for each participant.
In Figure \ref{fig:imuposthigh}, we show the results of the IMU positioning optimization for all participants' thigh IMUs. We use the trochanter and knee markers as reference for the hip and knee joints. We present the manually measured position of the thigh IMU in the dataset, which \cite{dorschkyLowerbodyInertialSensor2024} assumed was located on the segment axis. We show that we are able to recover this misplacement from the dataset. For most participants, the position estimate is on or very close to the IMU housing. To our knowledge, current methods can only estimate the distance of an IMU from the joint center, but not the distance of the IMU to the segment axis \cite{seelIMUBasedJointAngle2014}.
This discrepancy between the positioning from the dataset and our estimation was only apparent for the thigh IMUs, and the margin of improvement in the metrics after personalization is \change{very small}{negligible} \remove{(see Appendix~[])}. However, the personalization of the IMU positionings can make the model more robust to misplacements and misalignments when donning the IMUs.

\section{Discussion}
In this work, we presented SSPINNpose, a real-time method for the estimation of human movement dynamics from inertial sensor data that does not require labeled training data. Instead, it relies on self-supervision and physics information to find plausible motions. We show that SSPINNpose can accurately estimate joint angles, torques, and GRFs from IMU data, while outperforming state-of-the-art methods in terms of horizontal speed estimation. Additionally, SSPINNpose effectively identifies movement patterns from sparse IMU configurations and personalizes IMU placement on the body. Given its capability to work with minimal IMU configurations and allow for personalization, SSPINNpose is a promising approach for long-term monitoring of athletes and understanding injury mechanisms. 

\add{When comparing with the state of the art, it is important to note the operational differences. Current supervised learning methods are bound by the availability of labeled data, restricting them to laboratory environments. In contrast, optimal control-based methods do not require labeled data and can theoretically process any IMU recording, but are computationally expensive and infeasible for real-time inference. SSPINNpose bridges this gap by offering the best of both worlds, as it can be trained on arbitrary, unlabeled IMU data similar to optimal control methods, while offering real-time inference capabilities like supervised learning methods. Furthermore, these state-of-the-art methods typically rely on pre-segmented gait cycles for inference}~\cite{dorschkyEstimationGaitKinematics2019,dorschkyCNNBasedEstimationSagittal2020,dorschkyComparingSparseInertial2025}. \add{This segmentation also typically assumes symmetric and periodic motion, which limits generalization to arbitrary or non-cyclic movements. We argue that inference on continuous IMU data is a more realistic capability for real-world applications, avoiding the limitations imposed by gait cycle segmentation.}

\add{An important consideration is that only GRFs are directly measured, while joint angles and torques are estimated from OMC data and inherit systematic biases} \cite{fleischmannExploringDatasetBias2025,werlingRapidBilevelOptimization2022}\add{, which supervised methods replicate. SSPINNpose learns from physical principles and measured data, potentially avoiding such biases.}

In SSPINNpose, the network implicitly learns the interconnection between movement kinematics and dynamics. In contrast, other real-time capable methods, such as  \cite{yiPhysicalInertialPoser2022,shimadaPhysCapPhysicallyPlausible2020}, use \add{the }kinematic\remove{al} network output as an input signal for a PD controller and dynamics optimizer. Whether direct or two-stage inference is preferable depends on the application. Direct inference is faster and the dynamics are not subject to propagation errors from the kinematics. On the other hand, the two-stage inference can help with generalization towards unseen movements and can be more robust to noise. 
When testing SSPINNpose with PIP's PD controller and optimizer, we found that the the PD controller hyperparameters needed to be adjusted. Depending on the hyperparameters, the PD controller could be used to slightly reduce the JTE while raising the JAE and smooth\add{en}ing the motion (see Appendix \ref{app:additionalresults} for implementation details).

Our method contains a number of assumptions and simplifications. As in \add{PIP} \cite{yiPhysicalInertialPoser2022}, we assume that the ground is flat and the foot cannot slide. Information about the ground is present in the IMU data, and has been exposed in recent work by inferring terrain height maps \cite{jiangTransformerInertialPoser2022}. It has yet to be shown whether the terrain information can also be learned without labels.
The interaction between foot and ground is modeled as a linear spring-damper system. Furthermore, the multibody dynamics model is based on a generic template, which \change[3]{is due to a}{justified by a} lack of personalization options\add[3]{ that are based solely on IMU data. Recently, methods for in-the-wild musculoskeletal model personalization based on images and LiDAR have been proposed and successfully applied to IMU data}~\cite{mcconnochieOptimalControlSimulations2025a,gambietzSmartphoneImagesMusculoskeletal2025a}, \add[3]{but are not available with existing datasets.}
As we fit towards IMU signals that are noisy, our model can learn to replicate that noise and becomes less physically plausible, which we are mitigating, but not eliminating, by augmenting the input data with Gaussian noise. Our model is able to accurately estimate human movement dynamics despite these limitations, therefore we consider them to be an opportunity to make the estimations more accurate in the future. 

\add{We further evaluated the performance of SSPINNpose on sparse sensor configurations. We observed that the error metrics generally increased depending on the specific sensor placement.} Figure~\ref{fig:spawsestick} \add{shows that the estimated pelvis orientation differs depending on whether a pelvis IMU is included.}
Therefore, \add{we concluded that} there is likely a discrepancy between the actual, physically plausible, trunk orientation and the IMU orientation, i.e. the pelvis IMU might not be correctly aligned, which was also observed in \cite{dorschkyComparingSparseInertial2025}. 
\remove{Compared to [4], our increases in errors are similar for the F and FP configurations, but higher for the FT configuration. We believe our method is more affected by soft tissue artefacts, measurement errors caused by the movement of skin and muscle, from thigh IMUs compared to the optimal control method. As there is no hard constraint in SSPINNpose, it can trade off physical correctness for a better fit to the IMU signals, especially when they contain noise. On the other hand, hard constraints in optimal control do not allow physically incorrect motions. The pelvis and foot IMUs, however, are less affected by soft-tissue artefacts.}

\add{Finally, we showed that SSPINNpose can reconstruct misplacements of the IMUs on the body}. 
\add{
While optimizing for the IMU positions, we can also optimize for other input parameters, such as body segment parameters and ground contact model parameters.} However, \change{there is no validation for the correctness of body constants and ground contact model parameters on the given dataset}{we cannot validate these parameters for the datasets used in this paper,}, as that would require medical imaging. Thus, we excluded these parameters \change{from the IMU positioning optimization}{and only optimized for IMU positioning, which could be validated}. However, when we optimized the body constants, we found that only the moments of inertia yielded unrealistic values, as they converged to zero. The convergence to zero is due to the formulation of Kane's loss, which favours \add{lower} moments of inertia, as they lead to less force and therefore less physics error in general. For the body weight, the same issue would apply, but we mitigated that by optimizing for the weight distribution instead of the body weight itself. 

There is room for more detailed modeling. For example, the model could be extended to 3D. \change{As current 3D optimal control simulations}{While complex 3D movements such as cutting} have only been successfully solved for simulated IMU data, \change[2]{however, we are unsure whether SSPINNpose would be able to solve these tasks.}{gait and running have been simulated from real IMU data} \cite{mcconnochieOptimalControlSimulations2025a} \add[2]{, which makes us confident that SSPINNpose can also be extended to 3D.}
To make the model even more detailed and biomechanically accurate, the torque actuators should be replaced with muscles. Furthermore, an unsupervised method for terrain mapping could be integrated. For supervised inertial pose estimation systems, \change{this}{terrain mapping} has already been shown to be possible \cite{jiangTransformerInertialPoser2022}. Additionally, other data sources such as optical markers or radar could be tested. 

Predictive simulations, an optimal control task where no reference data is given, are also of interest for future work. Concretely, the IMU tracking term would be replaced with a physiological movement objective, such as metabolic cost minimization. Here, self-supervised learning can be used to obtain a controller. The goal for SSPINNpose in this setting would be to be faster than model predictive control methods while being more accurate than reinforcement learning methods. Especially in muscle-driven simulations, the differentiable physics model is advantageous over reinforcement learning, as exploration for co-actuated systems is difficult \cite{schumacherDEPRLEmbodiedExploration2023}. 

In conclusion, SSPINNpose contributes to human movement analysis by providing a real-time method for the estimation of human movement dynamics from inertial sensor data without the need for labeled training data, which is crucial for movement analysis in the natural environment.

%

\section*{Acknowledgments}
  This work was funded by the Deutsche Forschungsgemeinschaft (DFG, German Research Foundation) – SFB
  1483 – Project-ID 442419336, EmpkinS

\textbf{Author contributions:} Conceptualization: MG, ED, AK; Methodology, Software, Investigation: MG, AA, MS, ED; Validation: MG; Writing - Original Draft, Review \& Editing: MG, ED, JM, AK; Supervision: ED, JM, AK.

\bibliography{iclr2025_conference}

@misc{bertichePBNSPhysicallyBased2021,
  title = {{{PBNS}}: {{Physically Based Neural Simulator}} for {{Unsupervised Garment Pose Space Deformation}}},
  shorttitle = {{{PBNS}}},
  author = {Bertiche, Hugo and Madadi, Meysam and Escalera, Sergio},
  year = 2021,
  month = may,
  number = {arXiv:2012.11310},
  eprint = {2012.11310},
  primaryclass = {cs},
  publisher = {arXiv},
  doi = {10.48550/arXiv.2012.11310},
  urldate = {2024-11-27},
  archiveprefix = {arXiv}
}

@article{chenDeterminingMotionsIMU2020,
  title = {Determining Motions with an {{IMU}} during Level Walking and Slope and Stair Walking},
  author = {Chen, Wei-Han and Lee, Yin-Shin and Yang, Ching-Jui and Chang, Su-Yu and Shih, Yo and Sui, Jien-De and Chang, Tian-Sheuan and Shiang, Tzyy-Yuang},
  year = 2020,
  month = jan,
  journal = {Journal of Sports Sciences},
  volume = {38},
  number = {1},
  pages = {62--69},
  issn = {0264-0414, 1466-447X},
  doi = {10.1080/02640414.2019.1680083},
  urldate = {2024-11-27},
  langid = {english}
}

@article{dorschkyCNNBasedEstimationSagittal2020,
  title = {{{CNN-Based Estimation}} of {{Sagittal Plane Walking}} and {{Running Biomechanics From Measured}} and {{Simulated Inertial Sensor Data}}},
  author = {Dorschky, Eva and Nitschke, Marlies and Martindale, Christine F and {van den Bogert}, Antonie J. and Koelewijn, Anne D. and Eskofier, Bjoern M.},
  year = 2020,
  journal = {Frontiers in Bioengineering and Biotechnology},
  volume = {8},
  number = {June},
  pages = {1--14}
}

@article{dorschkyComparingSparseInertial2025,
  title = {Comparing Sparse Inertial Sensor Setups for Sagittal-Plane Walking and Running Reconstructions},
  author = {Dorschky, Eva and Nitschke, Marlies and Mayer, Matthias and Weygers, Ive and Gassner, Heiko and Seel, Thomas and Eskofier, Bjoern M. and Koelewijn, Anne D.},
  year = 2025,
  month = feb,
  journal = {Frontiers in Bioengineering and Biotechnology},
  volume = {13},
  pages = {1507162},
  issn = {2296-4185},
  doi = {10.3389/fbioe.2025.1507162},
  urldate = {2025-03-10}
}

@article{dorschkyEstimationGaitKinematics2019,
  title = {Estimation of Gait Kinematics and Kinetics from Inertial Sensor Data Using Optimal Control of Musculoskeletal Models},
  author = {Dorschky, Eva and Nitschke, Marlies and Seifer, Ann-Kristin and Van Den Bogert, Antonie J. and Eskofier, Bjoern M.},
  year = 2019,
  month = oct,
  journal = {Journal of Biomechanics},
  volume = {95},
  pages = {109278},
  issn = {00219290},
  doi = {10.1016/j.jbiomech.2019.07.022},
  urldate = {2024-08-08},
  langid = {english}
}

@misc{dorschkyLowerbodyInertialSensor2024,
  title = {Lower-Body {{Inertial Sensor}} and {{Optical Motion Capture Recordings}} of {{Walking}} and {{Running}}},
  author = {Dorschky, Eva and Nitschke, Marlies and Seifer, Ann-Kristin and {van den Bogert}, Antonie and Koelewijn, Anne and Eskofier, Bjoern},
  year = 2024,
  month = jun,
  publisher = {Zenodo},
  doi = {10.5281/ZENODO.11522050},
  urldate = {2024-08-26},
  copyright = {Creative Commons Attribution 4.0 International}
}

@article{fleischmannExploringDatasetBias2025,
  title = {Exploring {{Dataset Bias}} and {{Scaling Techniques}} in {{Multi-Source Gait Biomechanics}}: {{An Explainable Machine Learning Approach}}},
  shorttitle = {Exploring {{Dataset Bias}} and {{Scaling Techniques}} in {{Multi-Source Gait Biomechanics}}},
  author = {Fleischmann, Sophie and Dietz, Simon and Shanbhag, Julian and Wuensch, Annika and Nitschke, Marlies and Miehling, J{\"o}rg and Wartzack, Sandro and Leyendecker, Sigrid and Eskofier, Bjoern M. and Koelewijn, Anne D.},
  year = 2025,
  month = feb,
  journal = {ACM Transactions on Intelligent Systems and Technology},
  volume = {16},
  number = {1},
  pages = {1--19},
  issn = {2157-6904, 2157-6912},
  doi = {10.1145/3702646},
  urldate = {2025-03-10},
  langid = {english}
}

@misc{gambietzSmartphoneImagesMusculoskeletal2025a,
  title = {From {{Smartphone Images}} to {{Musculoskeletal Models}}: {{Personalized Inertial Parameter Estimation}}},
  shorttitle = {From {{Smartphone Images}} to {{Musculoskeletal Models}}},
  author = {Gambietz, Markus and Azam, Putri Qistina and Amon, Philipp and Wechsler, Iris and Hille, Eva Maria and Menzel, Timo and Ott, Tabea and Botsch, Mario and Braun, Matthias and Miehling, J{\"o}rg and McMahon, Katie L. and Koelewijn, Anne D.},
  year = 2025,
  month = jul,
  publisher = {Bioengineering},
  doi = {10.1101/2025.07.08.663673},
  urldate = {2026-02-02},
  archiveprefix = {Bioengineering},
  copyright = {http://creativecommons.org/licenses/by/4.0/},
  langid = {english}
}

@article{heiderscheitIdentifyingTimeOccurrence2005,
  title = {Identifying the Time of Occurrence of a Hamstring Strain Injury during Treadmill Running: {{A}} Case Study},
  shorttitle = {Identifying the Time of Occurrence of a Hamstring Strain Injury during Treadmill Running},
  author = {Heiderscheit, Bryan C. and Hoerth, Dina M. and Chumanov, Elizabeth S. and Swanson, Stephen C. and Thelen, Brian J. and Thelen, Darryl G.},
  year = 2005,
  month = dec,
  journal = {Clinical Biomechanics},
  volume = {20},
  number = {10},
  pages = {1072--1078},
  issn = {02680033},
  doi = {10.1016/j.clinbiomech.2005.07.005},
  urldate = {2024-08-28},
  copyright = {https://www.elsevier.com/tdm/userlicense/1.0/},
  langid = {english}
}

@article{hernandezLowerBodyKinematics2021,
  title = {Lower Body Kinematics Estimation from Wearable Sensors for Walking and Running: {{A}} Deep Learning Approach},
  shorttitle = {Lower Body Kinematics Estimation from Wearable Sensors for Walking and Running},
  author = {Hernandez, Vincent and Dadkhah, Davood and Babakeshizadeh, Vahid and Kuli{\'c}, Dana},
  year = 2021,
  month = jan,
  journal = {Gait \& Posture},
  volume = {83},
  pages = {185--193},
  issn = {09666362},
  doi = {10.1016/j.gaitpost.2020.10.026},
  urldate = {2024-11-27},
  langid = {english}
}

@article{hochreiter1997long,
  title = {Long Short-Term Memory},
  author = {Hochreiter, S},
  year = 1997,
  journal = {Neural Computation MIT-Press}
}

@article{huangDeepInertialPoser2018,
  title = {Deep {{Inertial Poser}}: {{Learning}} to {{Reconstruct Human Pose}} from {{Sparse Inertial Measurements}} in {{Real Time}}},
  author = {Huang, Yinghao and Kaufmann, Manuel and Aksan, Emre and Black, Michael J. and Hilliges, Otmar and {Pons-Moll}, Gerard},
  year = 2018,
  month = nov,
  journal = {ACM Transactions on Graphics, (Proc. SIGGRAPH Asia)},
  volume = {37},
  pages = {185:1-185:15},
  publisher = {ACM},
  doi = {10.1145/3272127.3275108},
  note = {Two first authors contributed equally}
}

@article{jangPhysicsguidedSelfsupervisedLearning2024,
  title = {Physics-guided Self-supervised Learning: {{Demonstration}} for Generalized {{RF}} Pulse Design},
  shorttitle = {Physics-guided Self-supervised Learning},
  author = {Jang, Albert and He, Xingxin and Liu, Fang},
  year = 2024,
  month = oct,
  journal = {Magnetic Resonance in Medicine},
  pages = {mrm.30307},
  issn = {0740-3194, 1522-2594},
  doi = {10.1002/mrm.30307},
  urldate = {2024-11-27},
  langid = {english}
}

@inproceedings{jiangTransformerInertialPoser2022,
  title = {Transformer {{Inertial Poser}}: {{Real-Time Human Motion Reconstruction}} from {{Sparse IMUs}} with {{Simultaneous Terrain Generation}}},
  booktitle = {{{SIGGRAPH Asia}} 2022 {{Conference Papers}}},
  author = {Jiang, Yifeng and Ye, Yuting and Gopinath, Deepak and Won, Jungdam and Winkler, Alexander W. and Liu, C. Karen},
  year = 2022,
  series = {{{SA}} '22},
  publisher = {Association for Computing Machinery},
  address = {Daegu, Republic of Korea},
  doi = {10.1145/3550469.3555428},
  isbn = {978-1-4503-9470-3}
}

@book{kaneDynamicsTheoryApplications1985,
  title = {Dynamics: Theory and Applications},
  shorttitle = {Dynamics},
  author = {Kane, Thomas R. and Levinson, David A.},
  year = 1985,
  series = {{{McGraw Hill}} Series in Mechanical Engineering},
  publisher = {McGraw-Hill},
  address = {New York, NY},
  isbn = {978-0-07-037846-9},
  langid = {english}
}

@article{karatsidisMusculoskeletalModelbasedInverse2019,
  title = {Musculoskeletal Model-Based Inverse Dynamic Analysis under Ambulatory Conditions Using Inertial Motion Capture},
  author = {Karatsidis, Angelos and Jung, Moonki and Schepers, H Martin and Bellusci, Giovanni and Zee, Mark De and Veltink, Peter H and Skipper, Michael},
  year = 2019,
  journal = {Medical Engineering and Physics},
  volume = {65}
}

@inproceedings{Kim_2025_ICCV,
  title = {Probabilistic Inertial Poser ({{ProbIP}}): {{Uncertainty-aware}} Human Motion Modeling from Sparse Inertial Sensors},
  booktitle = {Proceedings of the {{IEEE}}/{{CVF}} International Conference on Computer Vision ({{ICCV}})},
  author = {Kim, Min and Jeon, Younho and Jo, Sungho},
  year = 2025,
  month = oct,
  pages = {25893--25902}
}

@misc{kingmaAdamMethodStochastic2017,
  title = {Adam: {{A Method}} for {{Stochastic Optimization}}},
  shorttitle = {Adam},
  author = {Kingma, Diederik P. and Ba, Jimmy},
  year = 2014,
  month = dec,
  number = {arXiv:1412.6980},
  eprint = {1412.6980},
  primaryclass = {cs},
  publisher = {arXiv},
  urldate = {2024-09-03},
  archiveprefix = {arXiv},
  note = {Comment: Published as a conference paper at the 3rd International Conference for Learning Representations, San Diego, 2015}
}

@misc{kocabasPACEHumanCamera2023,
  title = {{{PACE}}: {{Human}} and {{Camera Motion Estimation}} from in-the-Wild {{Videos}}},
  shorttitle = {{{PACE}}},
  author = {Kocabas, Muhammed and Yuan, Ye and Molchanov, Pavlo and Guo, Yunrong and Black, Michael J. and Hilliges, Otmar and Kautz, Jan and Iqbal, Umar},
  year = 2023,
  month = oct,
  number = {arXiv:2310.13768},
  eprint = {2310.13768},
  primaryclass = {cs},
  publisher = {arXiv},
  doi = {10.48550/arXiv.2310.13768},
  urldate = {2024-11-27},
  archiveprefix = {arXiv},
  note = {Comment: 3DV 2024. Project page: https://nvlabs.github.io/PACE/}
}

@article{kuderleGaitmapOpenEcosystem2024,
  title = {Gaitmap---{{An Open Ecosystem}} for {{IMU-Based Human Gait Analysis}} and {{Algorithm Benchmarking}}},
  author = {K{\"u}derle, Arne and Ullrich, Martin and Roth, Nils and Ollenschl{\"a}ger, Malte and Ibrahim, Alzhraa A. and Moradi, Hamid and Richer, Robert and Seifer, Ann-Kristin and Z{\"u}rl, Matthias and S{\^i}mpetru, Raul C. and Herzer, Liv and Prossel, Dominik and Kluge, Felix and Eskofier, Bjoern M.},
  year = 2024,
  journal = {IEEE Open Journal of Engineering in Medicine and Biology},
  volume = {5},
  pages = {163--172},
  issn = {2644-1276},
  doi = {10.1109/OJEMB.2024.3356791},
  urldate = {2024-08-26},
  copyright = {https://creativecommons.org/licenses/by/4.0/legalcode}
}

@article{limPredictionLowerLimb2019,
  title = {Prediction of {{Lower Limb Kinetics}} and {{Kinematics}} during {{Walking}} by a {{Single IMU}} on the {{Lower Back Using Machine Learning}}},
  author = {Lim, Hyerim and Kim, Bumjoon and Park, Sukyung},
  year = 2019,
  month = dec,
  journal = {Sensors},
  volume = {20},
  number = {1},
  pages = {130},
  issn = {1424-8220},
  doi = {10.3390/s20010130},
  urldate = {2024-11-27},
  copyright = {https://creativecommons.org/licenses/by/4.0/},
  langid = {english}
}

@article{liReconstructingWalkingDynamics2021,
  title = {Reconstructing {{Walking Dynamics From Two Shank-Mounted Inertial Measurement Units}}},
  author = {Li, Tong and Wang, Lei and Yi, Jingang and Li, Qingguo and Liu, Tao},
  year = 2021,
  month = dec,
  journal = {IEEE/ASME Transactions on Mechatronics},
  volume = {26},
  number = {6},
  pages = {3040--3050},
  issn = {1083-4435, 1941-014X},
  doi = {10.1109/TMECH.2021.3051724},
  urldate = {2024-09-03},
  copyright = {https://ieeexplore.ieee.org/Xplorehelp/downloads/license-information/IEEE.html}
}

@article{loperSMPLSkinnedMultiPerson2015,
  title = {{{SMPL}}: {{A Skinned Multi-Person Linear Model}}},
  author = {Loper, Matthew and Mahmood, Naureen and Romero, Javier and {Pons-Moll}, Gerard and Black, Michael J.},
  year = 2015,
  month = nov,
  journal = {ACM Trans. Graph.},
  volume = {34},
  number = {6},
  publisher = {Association for Computing Machinery},
  address = {New York, NY, USA},
  issn = {0730-0301},
  doi = {10.1145/2816795.2818013}
}

@article{mcconnochieOptimalControlSimulations2025a,
  title = {Optimal Control Simulations Tracking Wearable Sensor Signals Provide Comparable Running Gait Kinematics to Marker-Based Motion Capture},
  author = {McConnochie, Grace and Fox, Aaron S. and Bellenger, Clint and Thewlis, Dominic},
  year = 2025,
  month = mar,
  journal = {PeerJ},
  volume = {13},
  pages = {e19035},
  issn = {2167-8359},
  doi = {10.7717/peerj.19035},
  urldate = {2026-02-02},
  copyright = {https://creativecommons.org/licenses/by/4.0/},
  langid = {english}
}

@article{mcginleyReliabilityThreedimensionalKinematic2009a,
  title = {The Reliability of Three-Dimensional Kinematic Gait Measurements: {{A}} Systematic Review},
  shorttitle = {The Reliability of Three-Dimensional Kinematic Gait Measurements},
  author = {McGinley, Jennifer L. and Baker, Richard and Wolfe, Rory and Morris, Meg E.},
  year = 2009,
  month = apr,
  journal = {Gait \& Posture},
  volume = {29},
  number = {3},
  pages = {360--369},
  issn = {09666362},
  doi = {10.1016/j.gaitpost.2008.09.003},
  urldate = {2024-08-28},
  copyright = {https://www.elsevier.com/tdm/userlicense/1.0/},
  langid = {english}
}

@article{meurerSymPySymbolicComputing2017,
  title = {{{SymPy}}: Symbolic Computing in {{Python}}},
  shorttitle = {{{SymPy}}},
  author = {Meurer, Aaron and Smith, Christopher P. and Paprocki, Mateusz and {\v C}ert{\'i}k, Ond{\v r}ej and Kirpichev, Sergey B. and Rocklin, Matthew and Kumar, Amit and Ivanov, Sergiu and Moore, Jason K. and Singh, Sartaj and Rathnayake, Thilina and Vig, Sean and Granger, Brian E. and Muller, Richard P. and Bonazzi, Francesco and Gupta, Harsh and Vats, Shivam and Johansson, Fredrik and Pedregosa, Fabian and Curry, Matthew J. and Terrel, Andy R. and Rou{\v c}ka, {\v S}t{\v e}p{\'a}n and Saboo, Ashutosh and Fernando, Isuru and Kulal, Sumith and Cimrman, Robert and Scopatz, Anthony},
  year = 2017,
  month = jan,
  journal = {PeerJ Computer Science},
  volume = {3},
  pages = {e103},
  issn = {2376-5992},
  doi = {10.7717/peerj-cs.103},
  urldate = {2024-08-26},
  copyright = {http://creativecommons.org/licenses/by/4.0/},
  langid = {english}
}

@article{mohammadimoghadam3DGaitAnalysis2024,
  title = {{{3D}} Gait Analysis in Children Using Wearable Sensors: Feasibility of Predicting Joint Kinematics and Kinetics with Personalized Machine Learning Models and Inertial Measurement Units},
  shorttitle = {{{3D}} Gait Analysis in Children Using Wearable Sensors},
  author = {Mohammadi Moghadam, Shima and Ortega Auriol, Pablo and Yeung, Ted and Choisne, Julie},
  year = 2024,
  month = mar,
  journal = {Frontiers in Bioengineering and Biotechnology},
  volume = {12},
  pages = {1372669},
  issn = {2296-4185},
  doi = {10.3389/fbioe.2024.1372669},
  urldate = {2024-11-27}
}

@misc{molinaroDatasetTaskAgnosticExoskeleton2024,
  title = {Dataset for {{Task-Agnostic Exoskeleton Control}} via {{Biological Joint Moment Estimation}}},
  author = {Molinaro, Dean and Scherpereel, Keaton and Schonhaut, Ethan and Evangelopoulos, Georgios and Shepherd, Max and Young, Aaron},
  year = 2024,
  eprint = {1853/75759},
  eprinttype = {hdl},
  publisher = {Georgia Institute of Technology},
  doi = {10.35090/GATECH/75759},
  urldate = {2026-01-27}
}

@article{molinaroTaskagnosticExoskeletonControl2024a,
  title = {Task-Agnostic Exoskeleton Control via Biological Joint Moment Estimation},
  author = {Molinaro, Dean D. and Scherpereel, Keaton L. and Schonhaut, Ethan B. and Evangelopoulos, Georgios and Shepherd, Max K. and Young, Aaron J.},
  year = 2024,
  month = nov,
  journal = {Nature},
  volume = {635},
  number = {8038},
  pages = {337--344},
  issn = {0028-0836, 1476-4687},
  doi = {10.1038/s41586-024-08157-7},
  urldate = {2026-01-27},
  langid = {english}
}

@misc{nitschke3DKinematicsKinetics2023,
  title = {{{3D}} Kinematics and Kinetics of Change of Direction Motions Reconstructed from Virtual Inertial Sensor Data through Optimal Control Simulation},
  author = {Nitschke, Marlies and Dorschky, Eva and Leyendecker, Sigrid and Eskofier, Bjoern M and Koelewijn, Anne D},
  year = 2023,
  publisher = {Zenodo},
  doi = {10.5281/zenodo.8183292}
}

@article{roetenbergXsensMVNFull2013,
  title = {Xsens {{MVN}}: {{Full 6DOF}} Human Motion Tracking Using Miniature Inertial Sensors},
  author = {Roetenberg, Daniel and Luinge, Henk and Slycke, Per and others},
  year = 2013,
  journal = {Xsens Motion Technologies BV, Tech. Rep},
  volume = {1},
  pages = {1--7},
  publisher = {Citeseer}
}

@misc{santestebanSNUGSelfSupervisedNeural2022,
  title = {{{SNUG}}: {{Self-Supervised Neural Dynamic Garments}}},
  shorttitle = {{{SNUG}}},
  author = {Santesteban, Igor and Otaduy, Miguel A. and Casas, Dan},
  year = 2022,
  month = apr,
  number = {arXiv:2204.02219},
  eprint = {2204.02219},
  primaryclass = {cs},
  publisher = {arXiv},
  doi = {10.48550/arXiv.2204.02219},
  urldate = {2024-11-27},
  archiveprefix = {arXiv},
  note = {Comment: CVPR 2022 (Oral). Project website: http://mslab.es/projects/SNUG/}
}

@incollection{schmidtkeSelfsupervised3DHuman2023,
  title = {Self-Supervised {{3D Human Pose Estimation}} in {{Static Video}} via {{Neural Rendering}}},
  booktitle = {Computer {{Vision}} -- {{ECCV}} 2022 {{Workshops}}},
  author = {Schmidtke, Luca and Hou, Benjamin and Vlontzos, Athanasios and Kainz, Bernhard},
  editor = {Karlinsky, Leonid and Michaeli, Tomer and Nishino, Ko},
  year = 2023,
  volume = {13803},
  pages = {704--713},
  publisher = {Springer Nature Switzerland},
  address = {Cham},
  doi = {10.1007/978-3-031-25066-8_42},
  urldate = {2024-11-27},
  isbn = {978-3-031-25065-1 978-3-031-25066-8},
  langid = {english}
}

@misc{schumacherDEPRLEmbodiedExploration2023,
  title = {{{DEP-RL}}: {{Embodied Exploration}} for {{Reinforcement Learning}} in {{Overactuated}} and {{Musculoskeletal Systems}}},
  shorttitle = {{{DEP-RL}}},
  author = {Schumacher, Pierre and H{\"a}ufle, Daniel and B{\"u}chler, Dieter and Schmitt, Syn and Martius, Georg},
  year = 2023,
  month = apr,
  number = {arXiv:2206.00484},
  eprint = {2206.00484},
  primaryclass = {cs},
  publisher = {arXiv},
  doi = {10.48550/arXiv.2206.00484},
  urldate = {2025-03-10},
  archiveprefix = {arXiv}
}

@article{seelIMUBasedJointAngle2014,
  title = {{{IMU-Based Joint Angle Measurement}} for {{Gait Analysis}}},
  author = {Seel, Thomas and Raisch, J{\"o}rg and Schauer, Thomas},
  year = 2014,
  month = apr,
  journal = {Sensors},
  volume = {14},
  number = {4},
  pages = {6891--6909},
  issn = {1424-8220},
  doi = {10.3390/s140406891},
  urldate = {2025-03-10},
  copyright = {https://creativecommons.org/licenses/by/3.0/},
  langid = {english}
}

@misc{shimadaPhysCapPhysicallyPlausible2020,
  title = {{{PhysCap}}: {{Physically Plausible Monocular 3D Motion Capture}} in {{Real Time}}},
  shorttitle = {{{PhysCap}}},
  author = {Shimada, Soshi and Golyanik, Vladislav and Xu, Weipeng and Theobalt, Christian},
  year = 2020,
  month = dec,
  number = {arXiv:2008.08880},
  eprint = {2008.08880},
  primaryclass = {cs},
  publisher = {arXiv},
  doi = {10.48550/arXiv.2008.08880},
  urldate = {2024-11-27},
  archiveprefix = {arXiv},
  note = {Comment: 16 pages, 11 figures}
}

@inproceedings{simoncolomarSmoothingZUPTaidedINSs2012,
  title = {Smoothing for {{ZUPT-aided INSs}}},
  booktitle = {2012 {{International Conference}} on {{Indoor Positioning}} and {{Indoor Navigation}} ({{IPIN}})},
  author = {Simon Colomar, David and Nilsson, John-Olof and Handel, Peter},
  year = 2012,
  month = nov,
  pages = {1--5},
  publisher = {IEEE},
  address = {Sydney, Australia},
  doi = {10.1109/IPIN.2012.6418869},
  urldate = {2024-08-27},
  isbn = {978-1-4673-1954-6 978-1-4673-1955-3}
}

@misc{solaQuaternionKinematicsErrorstate2017,
  title = {Quaternion Kinematics for the Error-State {{Kalman}} Filter},
  author = {Sol{\`a}, Joan},
  year = 2017,
  month = nov,
  number = {arXiv:1711.02508},
  eprint = {1711.02508},
  primaryclass = {cs},
  publisher = {arXiv},
  urldate = {2024-08-27},
  archiveprefix = {arXiv}
}

@article{tanSelfSupervisedLearningImproves2024,
  title = {Self-{{Supervised Learning Improves Accuracy}} and {{Data Efficiency}} for {{IMU-Based Ground Reaction Force Estimation}}},
  author = {Tan, Tian and Shull, Peter B. and Hicks, Jenifer L. and Uhlrich, Scott D. and Chaudhari, Akshay S.},
  year = 2024,
  month = jul,
  journal = {IEEE Transactions on Biomedical Engineering},
  volume = {71},
  number = {7},
  pages = {2095--2104},
  issn = {0018-9294, 1558-2531},
  doi = {10.1109/TBME.2024.3361888},
  urldate = {2024-08-28},
  copyright = {https://ieeexplore.ieee.org/Xplorehelp/downloads/license-information/IEEE.html}
}

@inproceedings{trumbleTotalCapture3D2017,
  title = {Total {{Capture}}: {{3D Human Pose Estimation Fusing Video}} and {{Inertial Sensors}}},
  booktitle = {2017 {{British Machine Vision Conference}} ({{BMVC}})},
  author = {Trumble, Matt and Gilbert, Andrew and Malleson, Charles and Hilton, Adrian and Collomosse, John},
  year = 2017
}

@article{vandenbogertImplicitMethodsEfficient2011a,
  title = {Implicit Methods for Efficient Musculoskeletal Simulation and Optimal Control},
  author = {Van Den Bogert, Antonie J. and Blana, Dimitra and Heinrich, Dieter},
  year = 2011,
  journal = {Procedia IUTAM},
  volume = {2},
  pages = {297--316},
  issn = {22109838},
  doi = {10.1016/j.piutam.2011.04.027},
  urldate = {2024-08-26},
  copyright = {https://www.elsevier.com/tdm/userlicense/1.0/},
  langid = {english}
}

@misc{vanwouweDiffusionPoserRealtimeHuman2024,
  title = {{{DiffusionPoser}}: {{Real-time Human Motion Reconstruction From Arbitrary Sparse Sensors Using Autoregressive Diffusion}}},
  shorttitle = {{{DiffusionPoser}}},
  author = {Van Wouwe, Tom and Lee, Seunghwan and Falisse, Antoine and Delp, Scott and Liu, C. Karen},
  year = 2024,
  month = mar,
  number = {arXiv:2308.16682},
  eprint = {2308.16682},
  primaryclass = {cs},
  publisher = {arXiv},
  urldate = {2024-08-26},
  archiveprefix = {arXiv},
  note = {Comment: accepted at CVPR2024}
}

@inproceedings{vonmarcardSparseInertialPoser2017,
  title = {Sparse Inertial Poser: {{Automatic}} 3d Human Pose Estimation from Sparse Imus},
  booktitle = {Computer Graphics Forum},
  author = {Von Marcard, Timo and Rosenhahn, Bodo and Black, Michael J and {Pons-Moll}, Gerard},
  year = 2017,
  volume = {36},
  pages = {349--360},
  publisher = {Wiley Online Library}
}

@article{wachterImplementationInteriorpointFilter2006,
  title = {On the Implementation of an Interior-Point Filter Line-Search Algorithm for Large-Scale Nonlinear Programming},
  author = {W{\"a}chter, Andreas and Biegler, Lorenz T},
  year = 2006,
  journal = {Mathematical Programming},
  volume = {106},
  number = {1},
  pages = {25--57},
  doi = {10.1007/s10107-004-0559-y}
}

@inproceedings{wallbankBicepsFemorisMuscle2024,
  title = {Biceps {{Femoris Muscle States}} Prior to and during a {{Hamstring Strain Injury}} Whilst {{Sprinting}}},
  booktitle = {{{ISBS Proceedings Archive}}},
  author = {Wallbank, Kevin and Ede, Carlie and Blenkinsop, Glen and Allen, Sam},
  year = 2024,
  volume = {42: Iss 1, Article 193},
  address = {Salzburg}
}

@article{wangEstimationLowerLimb2023,
  title = {Estimation of {{Lower Limb Joint Angles}} and {{Joint Moments}} during {{Different Locomotive Activities Using}} the {{Inertial Measurement Units}} and a {{Hybrid Deep Learning Model}}},
  author = {Wang, Fanjie and Liang, Wenqi and Afzal, Hafiz Muhammad Rehan and Fan, Ao and Li, Wenjiong and Dai, Xiaoqian and Liu, Shujuan and Hu, Yiwei and Li, Zhili and Yang, Pengfei},
  year = 2023,
  month = nov,
  journal = {Sensors},
  volume = {23},
  number = {22},
  pages = {9039},
  issn = {1424-8220},
  doi = {10.3390/s23229039},
  urldate = {2024-11-27},
  copyright = {https://creativecommons.org/licenses/by/4.0/},
  langid = {english}
}

@inproceedings{wanSelfSupervised3DHand2019,
  title = {Self-{{Supervised 3D Hand Pose Estimation Through Training}} by {{Fitting}}},
  booktitle = {2019 {{IEEE}}/{{CVF Conference}} on {{Computer Vision}} and {{Pattern Recognition}} ({{CVPR}})},
  author = {Wan, Chengde and Probst, Thomas and Van Gool, Luc and Yao, Angela},
  year = 2019,
  month = jun,
  pages = {10845--10854},
  publisher = {IEEE},
  address = {Long Beach, CA, USA},
  doi = {10.1109/CVPR.2019.01111},
  urldate = {2024-11-27},
  copyright = {https://ieeexplore.ieee.org/Xplorehelp/downloads/license-information/IEEE.html},
  isbn = {978-1-7281-3293-8}
}

@article{werlingRapidBilevelOptimization2022,
  title = {Rapid Bilevel Optimization to Concurrently Solve Musculoskeletal Scaling, Marker Registration, and Inverse Kinematic Problems for Human Motion Reconstruction},
  author = {Werling, Keenon and Raitor, Michael and Stingel, Jon and Hicks, Jennifer L and Collins, Steve and Delp, Scott and Liu, C Karen},
  year = 2022,
  journal = {bioRxiv},
  pages = {2022--08},
  publisher = {Cold Spring Harbor Laboratory}
}

@inproceedings{winklerQuestSimHumanMotion2022,
  title = {{{QuestSim}}: {{Human Motion Tracking}} from {{Sparse Sensors}} with {{Simulated Avatars}}},
  booktitle = {{{SIGGRAPH Asia}} 2022 {{Conference Papers}}},
  author = {Winkler, Alexander and Won, Jungdam and Ye, Yuting},
  year = 2022,
  series = {{{SA}} '22},
  publisher = {Association for Computing Machinery},
  address = {Daegu, Republic of Korea},
  doi = {10.1145/3550469.3555411},
  isbn = {978-1-4503-9470-3}
}

@book{winterBiomechanicsMotorControl2009a,
  title = {Biomechanics and Motor Control of Human Movement},
  author = {Winter, David A.},
  year = 2009,
  edition = {4th ed.},
  publisher = {Wiley},
  address = {Hoboken, N.J},
  doi = {10.1002/9780470549148},
  isbn = {978-0-470-39818-0}
}

@misc{xiePhysicsbasedHumanMotion2022a,
  title = {Physics-Based {{Human Motion Estimation}} and {{Synthesis}} from {{Videos}}},
  author = {Xie, Kevin and Wang, Tingwu and Iqbal, Umar and Guo, Yunrong and Fidler, Sanja and Shkurti, Florian},
  year = 2022,
  month = aug,
  number = {arXiv:2109.09913},
  eprint = {2109.09913},
  primaryclass = {cs},
  publisher = {arXiv},
  urldate = {2024-08-30},
  archiveprefix = {arXiv},
  note = {Comment: To appear in ICCV 2021}
}

@article{yiEgoLocateRealTimeMotion2023a,
  title = {{{EgoLocate}}: {{Real-Time Motion Capture}}, {{Localization}}, and {{Mapping}} with {{Sparse Body-Mounted Sensors}}},
  author = {Yi, Xinyu and Zhou, Yuxiao and Habermann, Marc and Golyanik, Vladislav and Pan, Shaohua and Theobalt, Christian and Xu, Feng},
  year = 2023,
  month = jul,
  journal = {ACM Trans. Graph.},
  volume = {42},
  number = {4},
  publisher = {Association for Computing Machinery},
  address = {New York, NY, USA},
  issn = {0730-0301},
  doi = {10.1145/3592099}
}

@article{yinShapeawareInertialPoser2025,
  title = {Shape-Aware {{Inertial Poser}}: {{Motion Tracking}} for {{Humans}} with {{Diverse Shapes Using Sparse Inertial Sensors}}},
  shorttitle = {Shape-Aware {{Inertial Poser}}},
  author = {Yin, Lu and Shi, Ziying and Wu, Yinghao and Yi, Xinyu and Xu, Feng and Guo, Shihui},
  year = 2025,
  month = dec,
  journal = {ACM Transactions on Graphics},
  volume = {44},
  number = {6},
  pages = {1--12},
  issn = {0730-0301, 1557-7368},
  doi = {10.1145/3763311},
  urldate = {2026-01-26},
  langid = {english}
}

@inproceedings{yiPhysicalInertialPoser2022,
  title = {Physical {{Inertial Poser}} ({{PIP}}): {{Physics-aware Real-time Human Motion Tracking}} from {{Sparse Inertial Sensors}}},
  booktitle = {{{IEEE}}/{{CVF Conference}} on {{Computer Vision}} and {{Pattern Recognition}} ({{CVPR}})},
  author = {Yi, Xinyu and Zhou, Yuxiao and Habermann, Marc and Shimada, Soshi and Golyanik, Vladislav and Theobalt, Christian and Xu, Feng},
  year = 2022,
  month = jun
}

@article{yiTransPoseRealTime3D2021,
  title = {{{TransPose}}: {{Real-Time 3D Human Translation}} and {{Pose Estimation}} with {{Six Inertial Sensors}}},
  author = {Yi, Xinyu and Zhou, Yuxiao and Xu, Feng},
  year = 2021,
  month = jul,
  journal = {ACM Trans. Graph.},
  volume = {40},
  number = {4},
  publisher = {Association for Computing Machinery},
  address = {New York, NY, USA},
  issn = {0730-0301},
  doi = {10.1145/3450626.3459786}
}

@inproceedings{yuanSimPoESimulatedCharacter2021,
  title = {{{SimPoE}}: {{Simulated Character Control}} for {{3D Human Pose Estimation}}},
  shorttitle = {{{SimPoE}}},
  booktitle = {2021 {{IEEE}}/{{CVF Conference}} on {{Computer Vision}} and {{Pattern Recognition}} ({{CVPR}})},
  author = {Yuan, Ye and Wei, Shih-En and Simon, Tomas and Kitani, Kris and Saragih, Jason},
  year = 2021,
  month = jun,
  pages = {7155--7165},
  publisher = {IEEE},
  address = {Nashville, TN, USA},
  doi = {10.1109/CVPR46437.2021.00708},
  urldate = {2024-11-27},
  copyright = {https://ieeexplore.ieee.org/Xplorehelp/downloads/license-information/IEEE.html},
  isbn = {978-1-6654-4509-2}
}

@misc{zhangDynamicInertialPoser2024,
  title = {Dynamic {{Inertial Poser}} ({{DynaIP}}): {{Part-Based Motion Dynamics Learning}} for {{Enhanced Human Pose Estimation}} with {{Sparse Inertial Sensors}}},
  shorttitle = {Dynamic {{Inertial Poser}} ({{DynaIP}})},
  author = {Zhang, Yu and Xia, Songpengcheng and Chu, Lei and Yang, Jiarui and Wu, Qi and Pei, Ling},
  year = 2024,
  month = mar,
  number = {arXiv:2312.02196},
  eprint = {2312.02196},
  primaryclass = {cs},
  publisher = {arXiv},
  urldate = {2024-09-13},
  archiveprefix = {arXiv},
  note = {Comment: Accepted by CVPR2024}
}

\appendix

\section{Implementation Details}\label{chap:implementationdetails}
\paragraph{RNN \& Hyperparameters:} We use a network architecture similar to physics inertial poser (PIP) \cite{yiPhysicalInertialPoser2022}. We use a LSTM with 2 layers with a hidden size of \change{256}{512}, while the output layers are of size 128 and 46, respectively. The LSTM has a dropout rate of \SI{40}{\percent}. Further hyperparameters, including the weighting between the loss terms, are listed in table \ref{tab:hyper}. We take the hyperparameters from PIP, as we use the same architecture. The loss weights were tuned manually. 
\begin{table}[h!]
   \caption{Hyperparameters in SSPINNpose.}
   \label{tab:hyper}
   \begin{center}
   \begin{tabular}{lc}
      \hline
      \textbf{Parameter} & \textbf{Value} \\
      \hline
      learning rate & $10^{-3}$ \\
      optimizer & Adam \\
      batch size & 32 \\
      criterion & MSE \\
      $\eta_{imu}$ & 0.25 \\
      $\lambda_K $ & 3.0 \\
      $\lambda_T $ & 3.0 \\
      $\lambda_{IMU} $ & 30.0 \\     
      $\lambda_{ankle} $ & 100.0 \\
      $\lambda_{B} $ & 10000.0 \\
      $\lambda_{\tau} $ & 1.0 \\
      $\lambda_{slide} $ & 30.0 \\
      $\lambda_{FS} $ & 1.0 \\
      \hline
   \end{tabular}
   \end{center}
\end{table}

\paragraph{Calculation of point kinematics:}
We list the equations to calculate the global kinematics, containing the positions ($x,y$) and angle $\alpha$, of a point $\vp=\{x,\dot{x},\ddot{x},y,\dot{y},\ddot{y},\alpha,\dot{\alpha},\ddot{\alpha}\}$, based on its parent segment, here. For calculation, the parent is defined by an offset $d_x, d_y$, a point $\vp'=\{x', \dot{x}',\ddot{x}',y',\dot{y}',\ddot{y}',\alpha',\dot{\alpha}',\ddot{\alpha}'\}$. First, $\{\alpha,\dot{\alpha},\ddot{\alpha}\}$ are set by adding the local coordinates $\vq_p$ to $p'$ for the respective point. Then, $\{x,\dot{x},\ddot{x},y,\dot{y},\ddot{y}\}$ are calculated as follows:

\begin{equation}
   x = x' + \cos(\alpha')d_x - \sin(\alpha')d_y,
\end{equation}
\begin{equation}
   y = y' + \sin(\alpha')d_x + \sin(\alpha')d_y,
\end{equation}
\begin{equation}
   \dot{x} = \dot{x}' - \left(\sin(\alpha')d_x + \cos(\alpha')d_y\right) \dot{\alpha'},
\end{equation}
\begin{equation}
   \dot{y} = \dot{y}' + \left(-\sin(\alpha')d_y + \cos(\alpha')d_x\right) \dot{\alpha'},
\end{equation}
\begin{equation}
   \ddot{x} = \ddot{x'} + \left(-d_x\dot{\alpha'}^2 - \ddot{\alpha'}d_y \right)  \cos{\alpha'} - \left(-d_y\dot{\alpha'}^2 + \ddot{\alpha'}d_x \right)  \sin{\alpha'},
\end{equation}
\begin{equation}
   \ddot{y} =  \ddot{y'} + \left(-d_x\dot{\alpha'}^2 - \ddot{\alpha'}d_y \right)  \sin{\alpha'} + \left(-d_y\dot{\alpha'}^2 + \ddot{\alpha'}d_x \right)  \cos{\alpha'}.
\end{equation}

The global kinematics are only directly estimated for the pelvis and the ankle. Therefore, the global kinematics based on the pelvis are first calculated for the hip joint position and pelvis IMU and then propagated along the kinematic chain. From the ankle kinematics that are seperately estimated, the heel and ankle point globals are calculated.

\paragraph{Ground contact model:}
We determine the ground contact point based on the global ankle rotation $\alpha_{ankle}$, where the contact point is positioned on the line between heel and toe. The exact position is determined as $(\tanh(\alpha_{ankle}*7)+1)/2$, where 1 corresponds to the toe and 0 to the heel. 
The GRF is calculated as: $\vF_y = - k\zeta(\beta \vp_{gc,y})\left(1-b\dot{\vp}_{gc,y}\right)/\beta $ with $\beta = 300$, stiffness $k=$ \SI{100}{{BW}\per\meter}, damping $b=$ \SI{0.75}{\newton\second\per\meter}, and $\vF_x = \mu_{max}\tanh(\hat{\rvmu}) \vF_y$, with $\mu_{max} = 0.5$. The global ankle kinematics $\tilde{\vp}_{ankle}$ are estimated seperately and supervised by the estimated forward kinematics of the ankle $\vp_{ankle}$: 
\begin{equation}
   \mathcal{L}_{GC} = \left(\frac{1}{n_{ankle}}\sum_{i=1}^{n_{ankle}} \left( \left( \tilde{\vp}_{ankle} - \vp_{ankle} \right) / \sigma(\vp_{ankle}) \right)\right)^2.
\end{equation}

\paragraph{Bounds on joint limits and maximum velocity:}
For hip and ankle, we set the joint ranges to $[-\pi/3,\pi/3]$. As the knee can extend less, its joint range was set to $[-\pi/3, 0.1]$. The maximum velocity was set to $[\SI{-10}{\meter\per\second}, \SI{10}{\meter\per\second}]$, while the vertical root position was set to $[\SI{0}{\meter},\SI{2}{\meter}]$.

\paragraph{Equations for the auxiliary losses:}
The torque minimization loss is calculated as:
\begin{equation}
   \mathcal{L}_{\tau} = \left(\sum \rvtau / max(\dot{\vp}_{0,x},1)\right)^2,
\end{equation}
where $\rvtau$ is the joint torque and $\dot{\vp}_{0,x}$ is the speed of the root translation in the sagittal plane. 
The sliding penalty loss is calculated as:
\begin{equation}
   \mathcal{L}_{slide} = \left(\frac{1}{n_{gc}}\sum_{i=1}^{n_{gc}} \left(  |\dot{\vp}_{gc,x}| \vF_{gc,y} \right)\right)^2,
\end{equation}
where $\dot{\vp}_{gc,x}$ is the horizontal speed of the foot and $\vF_{gc,y}$ is the vertical GRF.
The foot speed loss is calculated as:
\begin{equation}
   \mathcal{L}_{FS} = \left(\frac{1}{2}\sum_{\vp \in \vp_{ankle}}|\dot{\vp_x}-\dot{\vp}_{K,x}|-0.3max(\dot{\vp}_{K,x})\right)^2,
\end{equation}
where $\dot{\vp}_{K,x}$ is the reconstructed horizontal speed of the foot-worn IMU and $\dot{\vp}_x$ is the estimated horizontal speed of the foot-worn IMU.

\section{Comparison to 3D pose estimation}\label{app:3dpose}
Current state-of-the-art 3D pose estimation methods are typically evaluated on different metrics than those that biomechanists are interested in, which are listed in table \ref{tab:all_metricsCG}: \textit{1.) Jitter:} The third derivative of the joint positions in \unit{\kilo\meter\per\second\cubed}. \textit{2.) Global Orientation Error (GOE):} The mean absolute error (MAE) between estimated global segment orientations and those obtained from addBiomechanics, including the root orientation, in degrees. This term is similar to the SIP error, which measures the accuracy of global limb orientations in 3D. \textit{3.) Mean Absolute Joint Angle Error (JA-MAE):} The MAE between estimated joint angles and those obtained from addBiomechanics, including the root orientation, in degrees. \textit{4. Joint Positioning Error (JPE):} The mean distance between the knee and ankle position in our estimation and the position of the respective OMC marker, in \unit{\centi\meter}. The greater trochanter marker was aligned with the hip joint in our estimations.
\begin{table}[h!]
   \caption{The top half shows results of our baseline models on more additional metrics for walking and all movements. For comparison, results from PIP \cite{yiPhysicalInertialPoser2022} are listed in the bottom half on its datasets.}
   \label{tab:all_metricsCG}
   \begin{center}
   \begin{tabular}{lccccc}
      \hline
      \textbf{SSPINNpose} & \textbf{Jitter} & \textbf{GOE} & \textbf{JA-MAE} & \textbf{JPE} & \textbf{Latency} (\unit{\milli\second}) \\
      & [\unit{\kilo\meter\per\second}] &[deg] & [deg]&  [\unit{\centi\meter}]& [\unit{\milli\second}] \\
      \hline
      Walking & 0.67 & 4.8 & 6.7 & 6.8 & 3.5 \\
      All motions & 1.79 & 5.6 & 7.2 & 6.5 & 3.5 \\
      \hline
      \textbf{PIP (Dataset)} & & \textbf{SIP} & &\\
      && [deg] \\
      \hline
      DIP-IMU & 0.24 & 15.02 & 8.73 & 5.04 & 16 \\
      TotalCapture & 0.20 & 12.93 & 12.04 & 6.51 & 16 \\
      \hline
   \end{tabular}
   \end{center}
\end{table}

Compared to PIP \cite{yiPhysicalInertialPoser2022}, our results show lower angular errors, slightly higher positioning errors and higher jitter. None of these metrics is directly compareable due to different reasons:
\begin{itemize}
   \item Different model configuration: SMPL \cite{loperSMPLSkinnedMultiPerson2015} is a 3D model, which PIP used, that contains more joints and rotational degrees of freedom. Therefore, the rotational errors can be bigger, while the joint positions are closer to the reference data. The positioning of joints and their distances to the aligned root joint also influences the metrics. Jitter is affected similiarly as JPE.
   \item Different evaluation method in JPE: In state-of-the-art methods, the reference joint centers are found by fitting SMPL to the reference data. On the other hand, we believe that the sagittal position of the knee and ankle markers is more precisely reflecting the actual joint position. By this, our error contains propagates inaccuracies in scaling the multibody dynamics model and thus reveals IMU-driven model personalization as a new challenge.
   \item The datasets are different. Besides walking, DIP-IMU and TotalCapture contain gestures, freestyle and range of motion movements. Therefore, there is no fair comparison between our method and PIP. 
\end{itemize}

\section{Additional Results}\label{app:additionalresults}

\paragraph{Ablations:}\label{ablation}
To justify the importance of the individual loss terms and implementation details, we performed an ablation study. The results are shown in Table \ref{tab:abel}. The ablations are explained as follows: \remove{\textit{1.) w/o est-ankle:} We do not estimate ankle kinematics seperately, we use the full-body kinematics to estimate the GRFs instead.} \textit{1.) w/o input noise:} We remove the input noise from the IMU signals. \change{\textit{2.) w/o GRF minimum:} We remove the minimum bound on the GRFs.}{\textit{2.) w/o $\mathcal{L}_{B,\tau,slide,FS}$:} For each auxiliary loss, we remove it from the total loss and retrain the model from scratch. This includes the bounds, torque minimization, sliding penalty, and foot speed loss.} \textit{3.) w/ two contact points:} Instead of defining a single contact point based on the global foot angle, we set a fixed contact points for the foot and the heel, respectively. This is similar to the ground contact model in \cite{dorschkyEstimationGaitKinematics2019}.

We show that all ablations lead to a decrease in performance. We note that the \change{GRF minimum is}{bound terms $\mathcal{L}_B$ are} especially important because it prevents local minima where the model does not learn to interact with the ground. 

\begin{table}[h!]
   \caption{Quantitative results from the ablation study}
   \label{tab:abel}
   \begin{center}
   \begin{tabular}{lccccc}
      \hline
      \textbf{Model Version} & \textbf{JAE} & \textbf{JTE} & \textbf{GRF} & \textbf{Jitter} & \textbf{Speed} \\
      \hline
      Baseline & 9.1 & 3.8 & 16.6 & 1.79 & 0.15 \\
      w/o noise augmentation ($\eta_{imu}$ = \SI{0})& 10.8  & 6.1 & 36.1 & 1.87 & 0.28 \\
      w/o $\mathcal{L}_{B}$ & 13.1 & 7.0 & 27.6 & 2.08 & 0.95 \\
      w/o $\mathcal{L}_{\tau}$ & 11.7 & 6.3 & 25.4 & 1.82 & 0.55 \\
      w/o $\mathcal{L}_{slide}$ & 9.5 & 3.9 & 18.8 & 2.11 & 0.86 \\
      w/o $\mathcal{L}_{FS}$ & 9.4 & 3.9 & 14.9 & 1.66 & 0.28 \\
      w/ two contact points & 10.7 & 4.1 & 22.8 & 1.78 & 0.38 \\
      \hline
   \end{tabular}
   \end{center}
\end{table}

\paragraph{Sensitivity of loss weights:}\label{sens_loss_weights}
\add[3]{To test the sensitivity of the loss weights, we retrained SSPINNpose while doubling or halving the loss weights of each loss term individually. The results are shown in Table} \ref{tab:sens_loss_weights}. \add[3]{The results show that SSPINNpose is robust to variations in most loss terms, and the performance only degrades slightly. The highest sensitivity is observed when modifying the IMU loss weight $\mathcal{L}_{IMU}$ or the bounds loss weight $\mathcal{L}_{B}$, which can lead to a failure to learn meaningful motions. The results in Table} \ref{tab:sens_loss_weights} \add[3]{also indicate that there is slight variance in metrics between training runs.}

\begin{table}[h!]
   \caption{Sensitivity of loss weights. ''HIGH'' means that the respective loss weight was doubled, ''LOW'' means that the respective loss weight was halved.}
   \label{tab:sens_loss_weights}
   \begin{center}
   \begin{tabular}{llccccc}
      \hline
      \textbf{Weight Term} & \textbf{Mode} & \textbf{JAE} & \textbf{JTE} & 
      \textbf{GRF} & \textbf{Jitter} & \textbf{Speed} \\
      \hline
      & Baseline & 9.1 & 3.8 & 16.6 & 1.79 & 0.15 \\
      \hline
      $\mathcal{L}_{K}$ & LOW & 9.1 & 3.2 & 16.2 & 1.71 & 0.14 \\
                     & HIGH & 9.1 & 3.3 & 16.0 & 1.88 & 0.16 \\
      \hline
      $\mathcal{L}_{T}$ & LOW & 8.9 & 3.5 & 17.9 & 2.21 & 0.16 \\
                     & HIGH & 9.2 & 3.3 & 16.0 & 1.50 & 0.17 \\
      \hline
      $\mathcal{L}_{IMU}$ & LOW & 8.9 & 3.9 & 15.7 & 1.43 & 0.14 \\
                     & HIGH & 16.3 & 8.3 & 86.5 & 3.14 & 5.14 \\
      \hline
      $\mathcal{L}_{GC}$ & LOW & 9.6 & 3.9 & 15.4 & 1.56 & 0.21 \\
                     & HIGH & 9.3 & 3.0 & 15.2 & 1.76 & 0.30 \\
      \hline
      $\mathcal{L}_{B}$ & LOW & 11.7 & 3.1 & 16.7 & 1.73 & 0.39 \\
                     & HIGH & 11.6 & 5.2 & 20.5 & 1.70 & 0.54 \\
      \hline
      $\mathcal{L}_{\tau}$ & LOW & 8.9 & 5.7 & 18.5 & 1.86 & 0.15 \\
                     & HIGH & 9.8 & 3.6 & 15.7 & 1.73 & 0.27 \\
      \hline
      $\mathcal{L}_{slide}$ & LOW & 9.0 & 3.4 & 18.1 & 1.97 & 0.15 \\
                     & HIGH & 9.2 & 3.4 & 16.7 & 2.00 & 0.18 \\
      \hline
      $\mathcal{L}_{FS}$ & LOW & 9.2 & 3.9 & 15.1 & 1.50 & 0.15 \\
                     & HIGH & 9.3 & 3.1 & 15.4 & 1.90 & 0.20 \\
      \hline
   \end{tabular}
   \end{center}
\end{table}

\paragraph{SSPINNpose combined with PIP's second stage:}
To combine PIP with SSPINNposes second stage, we first estimate the global kinematics with SSPINNpose and track the joint angles $q$ and joint velocities $\dot{p}$ with a PD controller. Instead of projecting contact polygons from contact points, we deviated from the original PIP implementation by using the heel and toe positions as the contact points. We heuristically set the ground contact probablity as a function of the horizontal speed of the contact points and estimated GRFs. The PD controller was tuned manually ($k_{\lambda}$ = 0.01, $k_{res}$ = 0.1, $k_{\tau}$ = 0.1).

\remove{
\paragraph{Sparse IMU configurations:}}
\remove{In Figure} \ref{fig:spawsestick}\remove{, we show the stick figures for the sparse IMU configurations. We show that the model is able to estimate physically and visually plausible motions for all configurations. The errors are higher for the foot and thigh (FT) configuration, as the thigh IMUs are more prone to soft-tissue artefacts. The errors are lowest for the foot and pelvis (FP) configuration, as the pelvis IMU is less affected by soft-tissue artefacts.}

\remove{\paragraph{Sensitivity to IMU misplacement:}

We retrained SSPINNpose with IMU positions that were perturbed by varying offsets in the sagittal plane. We did a training run each with offsets of 2cm, 4cm, 6cm, 8cm, 10cm, 15cm, and 20cm in a random direction. Next, we tested whether SSPINNpose was able to recover either the IMU positions from the finetuning experiment in Section ... or the IMU positions in the dataset. The results in Table ... indicate that SSPINNpose is robust to small variations in IMU placement, but performance degrades with offsets of 10 cm or more. Manual IMU placement or measurement should usually be accurate within less than 10 cm, so we consider this a reasonable level of robustness. 
Recovering the IMU positions in the dataset is more challenging, but as long the model has been trained on reasonably accurate IMU positions, it can partially recover the IMU positions. This is shown by the optimized offsets compared to the finetuning experiment and dataset being smaller than the offsets used for training.

}

\section{Graphical Overview of SSPINNpose training and evaluation scheme} \label{app:headliner}
In Figure \ref{headliner}, we give an overview of the training and evaluation scheme of SSPINNpose. The explaination to the graphic is as follows:
\textbf{A:} We take continuous IMU signals, body constants, IMU positions and ground contact model parameters as input. \textbf{B:} We use a (Bi-) LSTM to output kinematics and joint torques. \textbf{C:} We show a stick figure of the estimated kinematics at $\{2.5, 3.0,...,4.5\}$ \unit{\second}. For two out of these frames, we also show the reference kinematics in grey. \textbf{D:} We supervise our model using the loss functions introduced in Section \ref{sec:sspinnpose}. Here we show: \textit{1.) Kane's Loss}, which has the same dimensionality as the multibody dynamics model's degrees of freedom. \textit{2.) Temporal Consistency Loss} for $\vp_{ankle,r}$, where the estimated velocity is shown in black and the estimated acceleration in red. The dashed lines represent the numerical differentiation of the position and velocity, respectively. \textit{3.) Virtual IMU:} The simulated IMU signals of a foot-worn IMU. \textit{4.) Foot-IMU speed:} The estimated speed of the foot-worn IMU, our model in blue and the kalman-filter based integration in green. The shaded area marks the zone where the speed error is zero. \textbf{E:} We show the biomechanical outcome variables. Dashed lines represent the reference data. \textit{1.) Kinematics:} Hip flexion: blue; knee flexion: red; ankle plantarflexion: green. \textit{2.)~Speed:} Translational velocity. \textit{3.)~Torques:} Knee flexion: red, ankle plantarflexion: green. The hip flexion torque is not shown as it is out of range, but it is not estimated correctly for this trial. \textit{4.) GRFs:} Vertical: blue, horizontal: red.
\begin{figure}[t]
   \begin{center}
   \input{figures/headliner_body.tex}
   \end{center}
   \caption{Overview of the SSPINNpose training and evaluation process. All data shown is from a single running bout at a max speed of \SI{4.9}{\meter\per\second}. The shaded area marks the time where the reference data was recorded.}
   \label{headliner}
\end{figure}

\end{document}